%% file: main.tex
\documentclass[10pt,twocolumn,letterpaper]{article}

\usepackage{cvpr}              

\usepackage{url}
\usepackage{amsthm}
\usepackage{bm}
\usepackage[dvipsnames]{xcolor}

\usepackage[most]{tcolorbox}
\usepackage{wrapfig}
\usepackage{multirow}
\usepackage{graphicx,xcolor,float}
\usepackage{threeparttable}
\usepackage[ruled,vlined]{algorithm2e}
\usepackage{colortbl}
\usepackage{color}
\usepackage{xcolor}
\usepackage[dvipsnames]{xcolor}
\usepackage{multirow}
\usepackage{tabularx}
\usepackage{float}
\usepackage{graphicx}
\usepackage{booktabs}
\usepackage{arydshln}
\usepackage{enumitem}
\usepackage{wrapfig}
\usepackage{caption}
\usepackage{graphicx}
\usepackage{makecell}
\usepackage{float} 
\usepackage{mathtools}

\usepackage{bbding}
\usepackage{subcaption}
\usepackage{makecell}
\usepackage{tabularx}
\usepackage{amssymb,mathrsfs,amsmath}
\usepackage{pifont}
\usepackage{xcolor}          %
\definecolor{blueToken}{RGB}{0,102,204}
\definecolor{redToken}{RGB}{210,34,34}
\usepackage{bbm}
\usepackage{fontawesome5}
\usepackage{placeins}

\usepackage{xcolor}
\usepackage[most]{tcolorbox}

\definecolor{promptbg}{RGB}{232, 240, 242}
\definecolor{promptframe}{RGB}{66, 117, 129}

\usepackage{amsmath}   %

\usepackage{xparse}    %

\NewDocumentCommand{\F}{m o g}{%
  \mathrm{#1}
  \IfValueT{#2}{_{\!#2}}
  \IfValueT{#3}{\!(#3)}
}

\SetCommentSty{mycommfont}
\SetKwComment{Comment}{$\triangleright$\ }{}

\usepackage{listings}

\usepackage{amssymb}

\input{preamble}
\definecolor{cvprblue}{rgb}{0.21,0.49,0.74}
\usepackage[pagebackref,breaklinks,colorlinks,allcolors=cvprblue]{hyperref}
\usepackage[capitalize]{cleveref}
\def\paperID{} 
\def\confName{CVPR}
\def\confYear{2026}

\title{Granulon: Awakening Pixel-Level Visual Encoders with Adaptive Multi-Granularity Semantics for MLLM}

\author{
Junyuan Mao$^{1*}$ \quad
Qiankun Li$^{2*}$ \quad
Linghao Meng$^{1}$ \quad
Zhicheng He$^{1}$ \\
Xinliang Zhou$^{2}$ \quad
Kun Wang$^{2}$ \quad
Yang Liu$^{2}$ \quad
Yueming Jin$^{1\dagger}$ \\
$^{1}$National University of Singapore \quad
$^{2}$Nanyang Technological University \\
{\tt\small junyuanm@u.nus.edu, qiankun.li@ntu.edu.sg, ymjin@nus.edu.sg} \\
{\small $^*$Equal contribution \quad $^\dagger$Corresponding authors}
}

\begin{document}
\maketitle
\input{sec/0_abstract}

\input{sec/1_intro}
\input{sec/2_related}

\input{sec/3_method}

\input{sec/4_exp}

\section{Conclusion}
We present \texttt{Granulon}, a \texttt{DINOv3}-based MLLM that adaptively modulates visual granularity under textual guidance. By pairing a text-conditioned Controller with AdaTA, \texttt{Granulon} converts granularity into a controllable semantic dimension, enabling dynamic trade-offs between fine-grained details and coarse-level abstraction.
Across various benchmarks, \texttt{Granulon} consistently surpasses \texttt{CLIP}- and \texttt{DINO}-based baselines in accuracy, robustness, and hallucination reduction. Unlike prior methods that refine semantic encoders, \texttt{Granulon} strengthens a pixel-level backbone with progressive, multi-scale abstraction, pointing toward a new direction for unifying low-level perception and high-level semantics in future MLLMs.

\section*{Acknowledgements}
This work was supported by Ministry of Education Tier 2 grant, Singapore (T2EP20224-0028) and Ministry of Education Tier 1 grant, Singapore (23-0651-P0001)

{
    \small
    \bibliographystyle{ieeenat_fullname}
    \bibliography{main}
}


\appendix
\newpage
\input{sec/X_suppl}

\end{document}

%% file: sec/0_abstract.tex
\begin{abstract}
Recent advances in multimodal large language models largely rely on \texttt{CLIP}-based visual encoders, which emphasize global semantic alignment but struggle with fine-grained visual understanding. In contrast, \texttt{DINOv3} provides strong pixel-level perception yet lacks coarse-grained semantic abstraction, leading to limited multi-granularity reasoning. To address this gap, we propose \texttt{Granulon}, a novel \texttt{DINOv3}-based MLLM with adaptive granularity augmentation. \texttt{Granulon} introduces a text-conditioned granularity Controller that dynamically adjusts the visual abstraction level according to the semantic scope of the textual input, and an Adaptive Token Aggregation module that performs granularity-guided pooling and relation-aware clustering to produce compact, semantically rich visual tokens. This design enables unified "pixel-to-fine-to-coarse" reasoning within a single forward pass. 
Extensive and interpretable experiments demonstrate that \texttt{Granulon} improves accuracy by \(\sim\!30\%\!\uparrow\) and reduces hallucination by \(\sim\!20\%\!\downarrow\), outperforming all visual encoders under identical settings. Code is available at \href{https://github.com/jinlab-imvr/Granulon}{Granulon}.
\vspace{-2 em}
\end{abstract}

%% file: sec/1_intro.tex
\section{Introduction}

Recent years have witnessed remarkable progress in multimodal large language models (MLLMs) \cite{yin2024survey, qin2025survey} across diverse tasks such as visual question answering \cite{kuang2025natural}, image captioning \cite{bucciarelli2024personalizing}, and multimodal reasoning \cite{kil2024mllm}.
A typical MLLM consists of two main components: a visual encoder and a language model, where the visual encoder extracts structured feature representations from input images \cite{fu2024mme}.
A high-quality visual encoder not only captures rich local textures and object details but also provides a robust representational basis for cross-modal reasoning \cite{li2025perception}.
This design enables MLLMs to achieve superior accuracy and generalization when interpreting complex visual scenes and answering various queries \cite{liang2024survey}.

Traditional visual encoders focus primarily on extracting pixel-level features for unimodal tasks such as classification \cite{kim2024vit}, segmentation \cite{wu2025medical}, or detection \cite{zhang2021vit}.
However, such representations lack semantic alignment and thus perform suboptimally in multimodal settings \cite{khan2022transformers}.
To address this limitation, many studies have built MLLMs upon \texttt{CLIP}-pretrained encoders \cite{hafner2021CLIP}, leveraging the powerful contrastive learning and cross-modal alignment capabilities of \texttt{CLIP} \cite{fan2023improving}.
The introduction of \texttt{CLIP} brings large-scale image–text alignment \cite{tu2025toward}, enabling MLLMs to perform well under zero-shot understanding and cross-domain scenarios \cite{sun2024alpha}.
Consequently, an increasing number of works explore enhanced \texttt{CLIP} variants, for instance, by introducing multi-scale feature integration or attention-based fusion to further improve cross-modal reasoning. Representative examples include \texttt{EVA-CLIP} \cite{sun2023eva}, which adopts stronger vision backbones for multi-level semantic consistency, and \texttt{SigLIP} \cite{tschannen2025SigLIP}, which employs a sigmoid-based similarity metric to enhance alignment stability and achieve substantial gains in captioning and VQA tasks.

\begin{figure*}[t]
\centering
\includegraphics[width=1.0\linewidth]{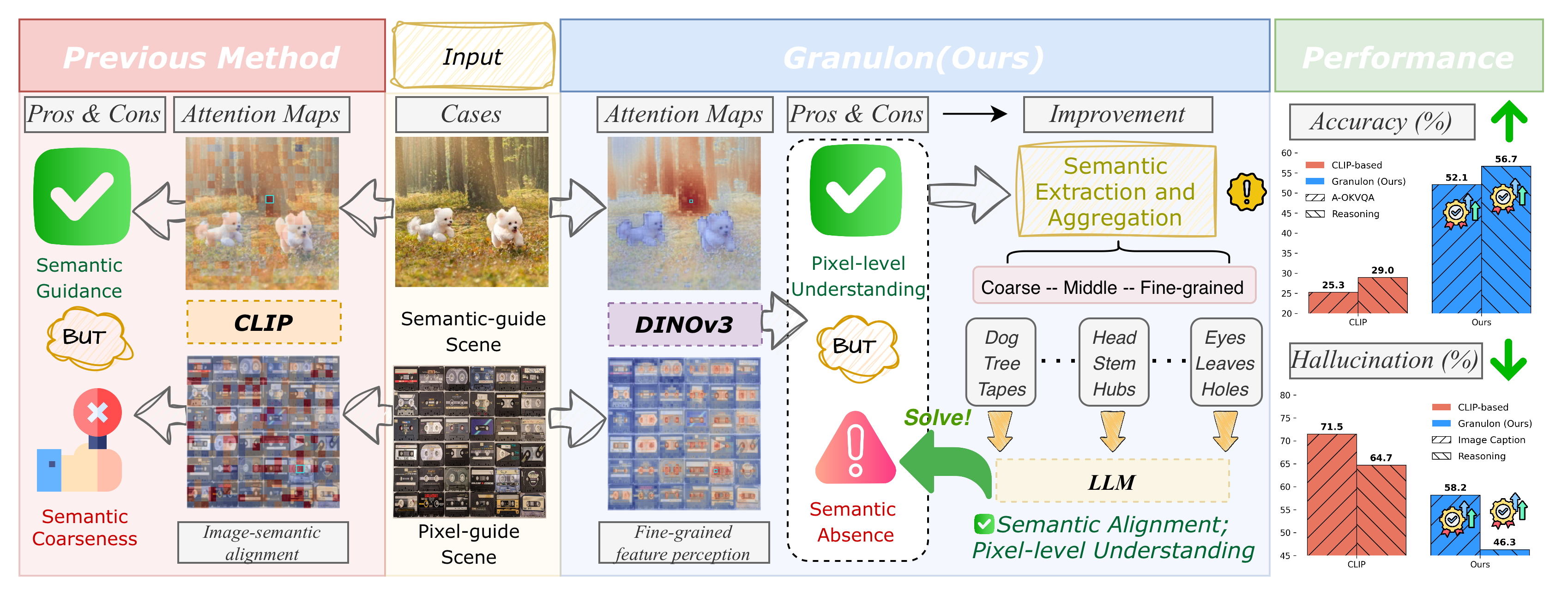}
\caption{\texttt{CLIP} tends to emphasize global semantics and \texttt{DINOv3} excels in pixel-level understanding. Our \textbf{\texttt{Granulon}} unleashes the potential of \texttt{DINOv3} with adaptive granularity augmentation, achieving consistent semantic alignment across different levels of detail and delivering superior multimodal reasoning performance.}
\label{fig:intro}
\vspace{-1.5 em}
\end{figure*}

However, despite the success of \texttt{CLIP} in semantic alignment and retrieval \cite{eslami2025mitigate, csizmadia2025distill}, its fixed-resolution global features still face limitations when handling tasks that require fine-grained visual understanding \cite{shen2021much}.
Because \texttt{CLIP} primarily learns semantic consistency through contrastive image–text pairs, it tends to emphasize global concepts while neglecting local textures or geometric details, leading to information loss and representation ambiguity in pixel-level understanding and reasoning \cite{zhou2024image}.
As a result, some researchers have shifted toward pixel-level self-supervised encoders to enhance fine-detail modeling.
In particular, \texttt{DINOv3} \cite{simeoni2025DINOv3}, with its self-distillation framework, exhibits outstanding capability in capturing relational structures and fine textures, and is widely regarded as one of the strongest detail-oriented visual encoders with inherent pixel reasoning ability.
Nevertheless, while \texttt{DINOv3} excels in fine-grained texture perception, it remains relatively limited in understanding coarse-grained structures and global semantics \cite{jiang2023CLIP}.
Some works combine \texttt{CLIP} and \texttt{DINO} to ease the gap between global semantics and fine details \cite{wysoczanska2024clip, zhang2025frozen}. However, such multi-encoder designs are computationally costly and do not address the core issue that a single encoder still lacks unified coarse-to-fine granularity in MLLMs.

As shown in Figure \ref{fig:intro}, to address these limitations, we move away from the prevailing CLIP-based paradigm, which offers strong semantic abstraction but lacks granularity flexibility, and instead revisit pixel-level encoders such as \texttt{DINOv3}.
Rather than relying on the fixed coarse semantic of CLIP, we introduce granularity augmentation as a structural mechanism that unlocks the underused semantic capacity of pixel-level encoders by injecting a controllable \textbf{\textit{pixel-fine-coarse}} semantic hierarchy into self-supervised representations.
This perspective opens a \textbf{\textit{previously underexplored angle}} for MLLMs: enabling pixel-level visual encoders to acquire task-adaptive semantic granularity that they intrinsically lack.

Building on this, we introduce \textbf{\texttt{Granulon}}, a DINOv3-based MLLM that can elevate granularity into a text-conditioned control dimension and integrate it with the pixel-level processing of \texttt{DINOv3}.
This differs from prior techniques that prune tokens based on bottom-up visual saliency or apply text-guided attention on fixed tokens, such as token pruning in DynamicViT \cite{rao2021dynamicvit} and EViT \cite{liang2022not}, and region-selection in LLaVA-NeXT \cite{li2024llava} and InternVL \cite{chen2024internvl}. Specifically, \texttt{Granulon} mainly comprises two key modules: \ding{182} a text-driven granularity Controller and \ding{183} an adaptive token aggregation (AdaTA) module, which create multi-granularity semantic tokens from pixels systematically.
The granularity Controller predicts the most suitable visual scale according to the linguistic complexity and referential scope of the input question, guiding subsequent semantic token aggregation in the visual stream.
Meanwhile, AdaTA performs granularity-guided pooling to adjust feature resolution, followed by relation-aware clustering that groups tokens with similar attention patterns, and finally applies quality-based selection to retain the most informative clusters. \textit{These semantic tokens compactly summarize both local textures and global semantics, enabling language model backbone to reason over multi-scale, semantic inputs for robust multimodal reasoning.}

We extensively evaluate \texttt{Granulon} across diverse benchmarks under \textbf{identical} experimental settings.
\texttt{Granulon} consistently outperforms \texttt{CLIP}- and \texttt{DINO}-based baselines, improving reasoning accuracy by \(\sim30\%\!\uparrow\) and reducing hallucination by \(\sim20\%\!\downarrow\). 
Ablations show that combining the Controller with AdaTA delivers up to \(10\%\) improvement while preserving token efficiency.
Our interpretability analysis also suggests that strengthening coarse-level abstraction in pixel-based visual encoders such as \texttt{DINOv3} represents a promising direction to enhance MLLM comprehension and output fidelity.

The main contributions of our work are as follows:
\begin{itemize}[leftmargin=2em,itemsep=-0.1em]
\item We identify a new direction for improving MLLMs, enhancing the coarse-grained abstraction capability of pixel-level visual encoders such as \texttt{DINOv3}.
\item We propose \textbf{\texttt{Granulon}}, which adaptively enhances visual feature granularity via a text-conditioned Controller and adaptive token aggregation.
\item Extensive and interpretable experiments demonstrate that \texttt{Granulon} not only improves accuracy but also brings a notable reduction in hallucination, outperforming all visual encoders under identical settings.

\end{itemize}

%% file: sec/2_related.tex
\section{Related Work}

\vspace{-0.3 em}
\paragraph{Semantic-level Visual Encoder.}
Visual encoders serve as the semantic bridge in MLLMs, transforming vision into representations that are discriminative and linguistically grounded \cite{bordes2024introduction}. 
While early approaches relied on \texttt{CLIP}-like architectures \cite{esfandiarpoor2024if, giahi2025vl} trained with large-scale image–text contrastive learning \cite{lee2022uniCLIP, yang2022vision, goel2022cyCLIP}, recent advances have shifted toward enhancing semantic abstraction and granularity of visual features. 
This trend aims to produce representations that better align with the rich semantic space required by large language models. 
Variants such as \texttt{EVA-CLIP} \cite{sun2023eva} and \texttt{SigLIP} \cite{zhai2023sigmoid} further improve representation fidelity by adopting stronger vision backbones and alternative similarity objectives based on sigmoid losses. 
Beyond purely contrastive encoders, hybrid frameworks like BLIP-2 \cite{li2023blip}, LLaVA \cite{schlarmann2024robust}, and QwenVL \cite{bai2025qwen2} couple frozen vision encoders with LLMs through dedicated bridging modules, enabling more effective multimodal alignment as well as instruction-following capabilities.

\vspace{-0.9 em}
\paragraph{Pixel-level Visual Encoder.}
Early masked image modeling frameworks such as BEiT \cite{bao2021beit} and MAE \cite{he2022masked} reconstruct masked patches to learn pixel-consistent embeddings that capture local texture and geometric structure. 
Subsequent variants, including iBOT \cite{zhou2021ibot} and MSN \cite{assran2022masked}, combine masked prediction with feature-level distillation to improve feature coherence. 
In parallel, self-distillation approaches like \texttt{DINO} \cite{caron2021emerging}, \texttt{\texttt{DINOv2}} \cite{oquab2023DINOv2, jose2025dinov2}, and \texttt{DINOv3} \cite{simeoni2025DINOv3} learn fine-grained relational features by enforcing cross-view consistency without any reconstruction target. 
These paradigms yield pixel-level encoders that excel in detailed understanding. 
However, these models often lack explicit mechanisms for semantic control, motivating our framework to bridge pixel- and semantic-level representations through unified multimodal fusion.

\vspace{-0.9 em}
\paragraph{Multimodel Large Language Models.}
With the rapid development of LLMs \cite{wang2025comprehensive, mao2025agentsafe, yu2025backdoor}, MLLMs have made remarkable progress in a wide range of multimodal applications \cite{zhang2024vision, li2025benchmark}, including visual question answering \cite{hartsock2024vision}, image captioning \cite{hu2022scaling}, and visual reasoning \cite{chen2024visual}. Early frameworks such as BLIP \cite{li2023blip} and Flamingo \cite{alayrac2022flamingo} demonstrated that coupling visual representations with language priors could enable powerful cross-modal reasoning.
Subsequent advances, notably LLaVA \cite{li2023llava}, leveraged GPT-4 to synthesize large-scale visual instruction datasets, which significantly improved performance in open-ended visual dialogue and reasoning tasks \cite{cocchi2025llava}. This paradigm inspired a series of instruction-tuned VLMs that further refined the alignment between vision and language modalities \cite{li2023vision, dai2023instructblip, doveh2024towards}.
Currently, open-source VLM families including Gemma3 \cite{team2025gemma}, QwenVL \cite{wang2024qwen2}, and InternVL \cite{chen2024internvl} have become the mainstream foundation models for multimodal understanding and instruction following.

%% file: sec/3_method.tex
\section{Methodology}
\begin{algorithm}[t]
\DontPrintSemicolon
\SetAlgoLined
\SetKwInOut{Input}{Input}\SetKwInOut{Output}{Output}
\Input{Dataset $\mathcal{D}=\{(\mathcal{I},\mathcal{T})\}$, vision encoder $\mathcal{V}_{vis}(\cdot)$, LLM $\mathcal{L}(\cdot)$, Controller $\pi_{\Theta}$}
\Output{Updated model parameters $\Theta$}

\ForEach{$(\mathcal{I},\mathcal{T}) \in \mathcal{D}$}{
  \tcc{1. Visual and textual encoding}
  $\bm{F} \leftarrow \mathcal{V}_{vis}(\mathcal{I}),
  \quad
  \mathcal{T}_e \leftarrow \mathsf{emb}_{l}(\mathcal{T})$ 
  
  \tcc{2. Granularity inference based on text}
  $g^{*} \leftarrow \Phi_{\mathrm{MLP}}\!\circ\!\Psi_{\mathrm{agg}}\!\circ\!\mathcal{L}^{(1)}(\mathcal{T}_e)$ 

  \tcc{3. Semantic feature modulation}
  $\mathcal{F}_{\text{mix}}(\mathcal{I}, \mathcal{T}) \leftarrow \Phi_{\gamma^{*}}\!\big[\bm{F} \oplus \mathcal{A}_{\pi_{\theta}}(\bm{F}; \mathcal{T}_e)\big] \oplus \mathcal{T}_e$
  
  \tcc{4. Multimodal reasoning}
  $\hat{y} \leftarrow \mathcal{L}(\bm{F}_{\text{mix}})$ 

  \tcc{5. Joint optimization}
  Compute pixel and semantic likelihoods: 
  $p_d=p_{\pi_{\Theta}}\!\big(C_{\text{pixel}}\!\mid\!\bm{F},\mathcal{T}\big)$, 
  $p_t=p_{\pi_{\Theta}}\!\big(C_{\text{semantic}}\!\mid\!\text{AdaTA}_{\pi_{\Theta}}(\bm{F}),\mathcal{T}\big)$\;
  $L = L_{task} -\big(\lambda_{d}\,\mathbb{E}_{\mathbf{v}_i}\log p_d 
  + \lambda_{t}\,\mathbb{E}_{\mathbf{t}_j}\log p_t\big)$\;
  Update $\Theta \leftarrow \Theta - \eta\nabla_{\Theta}L$\;
}
\caption{Workflow of \textbf{\texttt{Granulon}}}
\label{algo:Granulon}
\vspace{-0.2 em}
\end{algorithm}

\subsection{Overview}
\paragraph{Notations.}
Let the input be 
$\bm{X} = \{\mathcal{I}, \mathcal{T}\}$, where
$\mathcal{I}\in\mathbb{R}^{H\times W\times c_{in}}$ and $\mathcal{T}_e=\{t_{1},\dots ,t_{l}\} \in \mathbb{R}^{l \times D}$ tokenized questions of length $l$.
We denote the frozen vision encoder as $\mathcal{V}_{\text{vis}}$ and the language backbone as $\mathcal{L}$.
Feature encoders map images to latent maps
\(\bm{F}\in\mathbb{R}^{H'\times W'\times C'}\).
The granularity hypothesis space is formally defined as  
\begin{equation}
\pi_{\Theta}=\{g_{k}\}_{k=1}^{n},\quad g_{k}\triangleq\bigl(\alpha_{k},\beta_{k},\gamma_{k}\bigr)\in\mathbb{R}^{3},
\end{equation}
where $\alpha_{k}$ controls spatial down-sampling, $\beta_{k}$ regulates the cluster cardinality in token summarization, and $\gamma_{k}$ means the weights inside the multimodal projector.

\begin{figure*}[t]
\centering
\includegraphics[width=1.0\linewidth]{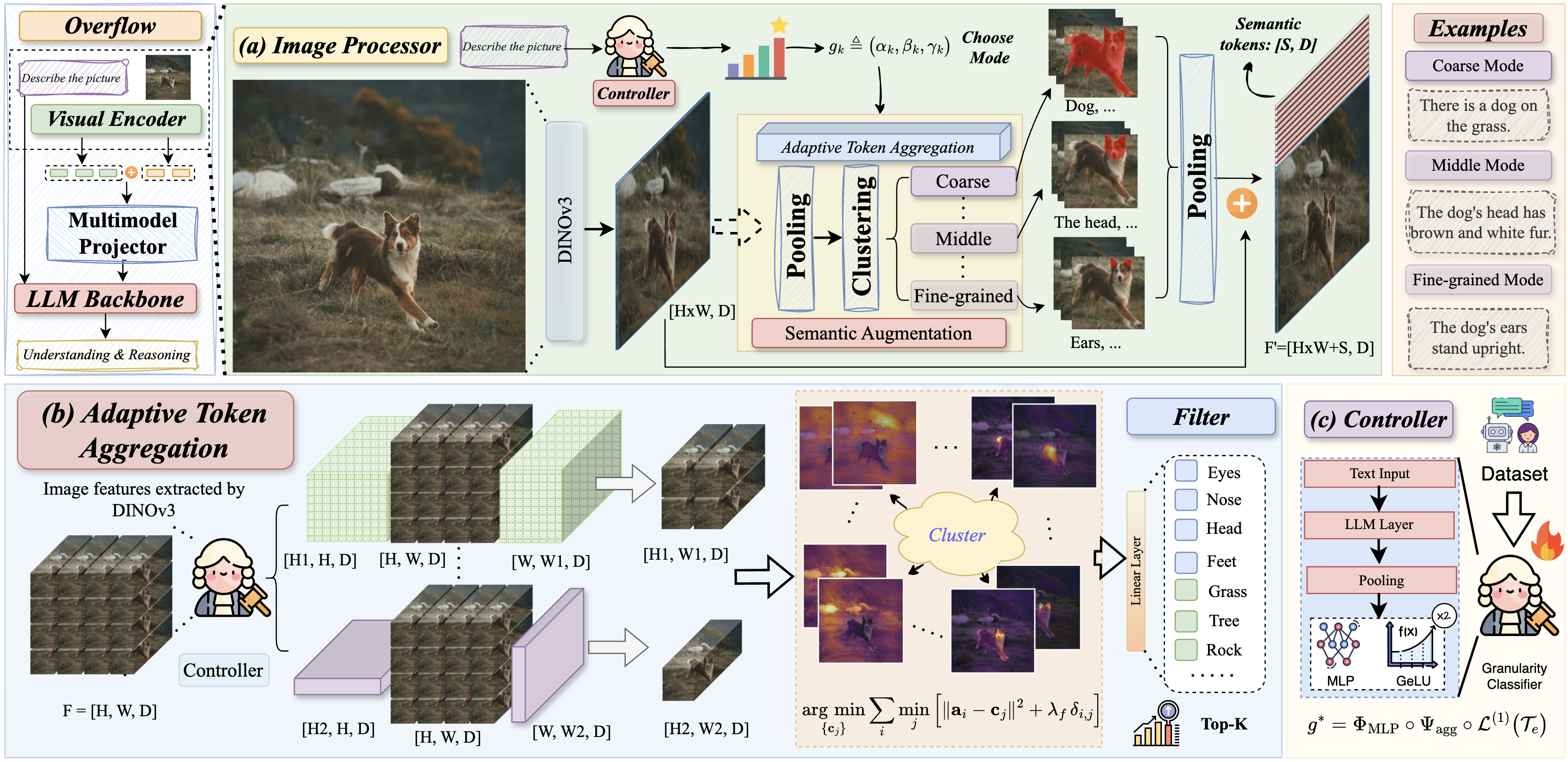}
\caption{Overview of $\textbf{\texttt{Granulon}}$. (a) The architecture of the image processor. (b) The detailed process of AdaTA that generates semantic tokens for multi-granularity augmentation. (c) The training and running of Controller.}
\label{fig:Granulon}
\vspace{-1.5 em}
\end{figure*}

\paragraph{Target.}
To encourage the model to exploit both pixel-level details and semantic-level summaries, we maximize the joint likelihood of two complementary token streams. The objective aggregates (i) per-patch confidence from image tokens and (ii) the cluster-level confidence from semantic tokens, regardless of the concrete granularity $g$:
\vspace{-0.3 em}
\begin{equation}
\begin{aligned}
\arg\max_{\pi_{\Theta}}\;
&= \mathbb{E}_{(\mathcal{I},\mathcal{T})\sim\mathcal{D}}
\Big[
\underbrace{\mathbb{E}_{\mathbf{v}_i\in\bm{F}}
\log p_{\pi_{\Theta}}\!\big(C_{\text{pixel}} \!\mid\! \mathbf{v}_i,\mathcal{T}\big)
}_{\text{\color{blueToken}\textbf{detail contribution}}}
\\[-2pt]
&+
\lambda\,
\underbrace{\mathbb{E}_{\mathbf{t}_j\in\mathcal{A}_{\pi_{\Theta}}(\bm{F})}
\log p_{\pi_{\Theta}}\!\big(C_{\text{sema}} \!\mid\! \mathbf{t}_j,\mathcal{T}\big)
}_{\text{\color{redToken}\textbf{granularity contribution}}}
\Big].
\end{aligned}
\label{eq:contrib_obj}
\end{equation}

For each token $\mathbf{v}_i\!\in\!\bm{F}$, 
$p_{\pi_{\Theta}}\!\big(C_{\text{pixel}}\!\mid\!\mathbf{v}_i,\mathcal{T}\big)$ 
quantifies the probability that $\mathbf{v}_i$ 
provides meaningful \emph{\color{blueToken}pixel-level contributions} to multimodal representations 
with context $\mathcal{T}$. 
$\mathcal{A}_{\pi_{\Theta}}(\bm{F})$ denotes the \emph{Adaptive Token Aggregation (AdaTA)} module 
parameterized by $\pi_{\Theta}$, 
which produces a compact set of \emph{\color{redToken}granular tokens} 
$\mathbf{t}_j$ that capture higher-level, structure-aware semantics. 
Accordingly, 
$p_{\pi_{\Theta}}\!\big(C_{\text{sema}}\!\mid\!\mathbf{t}_j,\mathcal{T}\big)$ 
measures the likelihood that a semantic token $\mathbf{t}_j$ 
contributes to the global semantic understanding or reasoning process. 
The balancing coefficient $\lambda$ controls the relative importance 
between local detail fidelity and global semantic granularity.

\paragraph{Training objective.}
Let $\hat{y}$ denote the LLM prediction and $y$ the ground-truth.
We use a standard task loss $L_{\text{task}}$ and add the contribution objective in Eq.~\eqref{eq:contrib_obj} as a regularizer that encourages informative pixel and semantic tokens under the question context.
The final loss is
\begin{equation}
L
=
L_{\text{task}}
+
\lambda_{d}\,L_{\text{pixel}}
+
\lambda_{t}\,L_{\text{sema}},
\end{equation}
where $L_{\text{pixel}}$ and $L_{\text{sema}}$ are the negative counterparts of the two expectations in Eq.~\eqref{eq:contrib_obj}, and $\lambda_{d}, \lambda_{t}$ balance their strengths.
This strategy makes \texttt{Granulon} learn to select an appropriate granularity and allocate contributions between pixel- and semantic-level tokens adaptive to tasks.

\paragraph{Overall Pipeline.}
As illustrated in Algorithm~\ref{algo:Granulon}, we propose \textbf{\texttt{Granulon}} which performs \emph{adaptive granularity modulation} to dynamically align visual abstraction with linguistic intent.  
\texttt{Granulon} can jointly learn to select the optimal visual granularity and to generate semantically coherent augmentation tokens, enabling fine-to-coarse reasoning. The process of \texttt{Granulon} is expressed as:
\begin{equation}
\label{eq:Granulon_pipeline}
\mathcal{F}_{\text{mix}}(\mathcal{I}, \mathcal{T})
=
\Phi_{\gamma^{*}}\!\big[\bm{F} \oplus \mathcal{A}_{\pi_{\theta}}(\bm{F}; \mathcal{T}_e)\big] \oplus \mathcal{T}_e,
\end{equation}

where
$\Phi_{\gamma^{*}}(\cdot)$ denotes the multimodal projector. $\mathcal{A}(\cdot)$ denotes the AdaTA process where $\mathcal{T}_e$ dynamically controls the aggregation mode according to the granularity of the textual input. Combined pixel- and semantic-level visual tokens with text embeddings in Figure \ref{fig:Granulon} \textit{(a)}, $\bm{F}_{\text{mix}}$ is fed into the LLM backbone, making \texttt{Granulon} achieve context-aware, multi-scale visual understanding while maintaining efficient and interpretable multimodal fusion.

\subsection{Granulon}
\texttt{Granulon} integrates a \texttt{DINOv3}-based vision encoder, a text-conditioned granularity Controller ($\triangleright$ Section \ref{sec:g2}), and a semantic token generation mechanism to achieve the fusion of dynamic, fine-to-coarse visual representation within a unified forward pass ($\triangleright$ Section \ref{sec:g3}).

\subsubsection{Text-conditioned Granularity Controller}
\label{sec:g2}

In Figure \ref{fig:Granulon} \textit{(c)}, We train a \textbf{granularity Controller} to dynamically predict the optimal visual abstraction level for each textual query.  
Given a question $\mathcal{T}$, the Controller outputs a granularity distribution $\bar{g}=\sum_{k=1}^{n} p_k g_k$, which modulates spatial pooling size and token cluster count through parameters $\alpha_k$ and $\beta_k$.  
This adaptive mechanism enables the model to adjust its perceptual resolution according to the linguistic complexity and referential scope of the input.

Formally, the Controller defines a differentiable mapping from text embeddings to the granularity space $\pi_{\theta}$:

\begin{equation}
\sum_{k=1}^{n} p(g_k \mid \mathcal{T}_e)\, g_k = \Phi_{\mathrm{MLP}}\circ\Psi_{\mathrm{agg}}\circ\mathcal{L}^{(1)}\big(\mathcal{T}_e\big),
\end{equation}
where
where $\mathcal{L}^{(1)}$ denotes the first block of LLM, 
$\Psi_{\mathrm{agg}}(\cdot)$ , is a pooling–projection to compact information.
and $\Phi_{\mathrm{MLP}}(\cdot)$ projects it to the final granularity parameters.

\noindent
\textbf{Interpretation.}
$\mathcal{L}^{(1)}$ serves as a linguistic encoder capturing surface dependencies and contextual emphasis within the query. 
Its output tokens are aggregated by $\Psi_{\text{agg}}$ through mean pooling and a non-linear projection:
\begin{equation}
\mathbf{h} = \mathbf{W}_{p}\,\sigma\big(\tfrac{1}{L}\sum_{i=1}^{L}\mathbf{E}^{(1)}_{i}\big),
\end{equation}
producing a compact textual descriptor $\mathbf{h}\in\mathbb{R}^{d_{c}}$.
This descriptor is then processed by the MLP head:
\begin{equation}
\Phi_{\text{MLP}}(\mathbf{h}) = \mathbf{W}_{2}\,\phi(\mathbf{W}_{1}\mathbf{h}+\mathbf{b}_{1})+\mathbf{b}_{2},
\end{equation}
yielding logits that are normalized into a categorical distribution 
$p(g\mid\mathcal{T})=\mathrm{softmax}(\Phi_{\text{MLP}}(\mathbf{h}))$,
where the final output $g^{*}=\{\alpha^{*}, \beta^{*}\}$ is obtained as $\arg\max_{g_{k}\in\bar{g}}p(g_{k}\mid\mathcal{T})$.

\noindent
\textbf{Supervision.}
We train the Controller using a text–granu-larity corpus annotated by GPT-4o~\cite{achiam2023gpt}, where questions are assigned $n$-dimensional weight vectors with their preferences across granularity scales.  
This supervision enables the Controller to learn mappings between linguistic intent and perceptual scope.  
For instance, in Figure \ref{fig:Granulon}, \emph{"What animals are in the image?"} favors a coarse configuration capturing global context, while \emph{"What color is the dog's ear?"} triggers a fine-grained mode focusing on local details.

\subsubsection{Adaptive Token Aggregation (AdaTA)}
\label{sec:g3}
Given that \texttt{DINOv3} excels at fine-grained texture modeling but is less sensitive to coarse structural cues, we introduce the \textbf{AdaTA} to enhance multi-scale perception and multimodal reasoning, as illustrated in Figure \ref{fig:Granulon} \textit{(b)}.  
AdaTA uses the granularity parameters predicted by Controller to produce \emph{semantic tokens} that capture visually salient regions at suitable granularity levels.  
The workflow includes three stages: (a) \emph{granularity-guided pooling}, (b) \emph{feature clustering}, and (c) \emph{feature refinement}.

\paragraph{(a) Granularity-guided Pooling.}
To modulate the visual resolution according to the output of Controller $\alpha^{*}$, we define a pooling operator parameterized by a kernel $\mathbf{K}_{\alpha^{*}}\in\mathbb{R}^{H'\times H'_{\alpha^{*}}}$ that performs spatial abstraction as
\begin{equation}
\mathbf{F}_{\alpha^{*}} = 
\mathbf{K}_{\alpha^{*}}^{\!\top}\,\mathbf{F}\,\mathbf{K}_{\alpha^{*}},
\quad
\mathbf{A}_{\alpha^{*}} = 
\mathbf{K}_{\alpha^{*}}^{\!\top}\,\mathbf{A}\,\mathbf{K}_{\alpha^{*}}.
\end{equation}
This operation jointly reduces the spatial dimensionality of both the feature and its attention field, preserving relative saliency while aligning token resolution with the desired granularity.
When $\alpha^{*}$ corresponds to a coarse setting, $\mathbf{K}_{\alpha^{*}}$ performs strong down-sampling (e.g., 4×4 pooling), whereas for fine settings it approaches the identity matrix.

\paragraph{(b) Feature Clustering.}
Given the pooled feature $\mathbf{F}_{\alpha^{*}}$ and its corresponding attention $\mathbf{A}_{\alpha^{*}}$, we derive the most representative visual prototypes via a mini-\textit{k}-means process.  
The output of the Controller
$\beta^{*}$ controls the cardinality of the cluster, which is formulated as:

\begin{equation}
\{\mathbf{c}_{j}\}_{j=1}^{M_{\beta^{*}}}
=
\operatorname*{arg\,min}_{\{\mathbf{c}_{j}\}}
\sum_{i}
\min_{j}
\Big[
\|\mathbf{a}_{i}-\mathbf{c}_{j}\|^{2}
+
\lambda_{f}\,
\delta_{i, j}\Big],
\end{equation}
where $\delta_{i, j}=\mathcal{D}_{\text{feat}}(\mathbf{F}_{\alpha^{*},i},\mathbf{F}_{\alpha^{*},j})$ measures the distance based on attention between the token $i$ and the group $j$.  
This joint spatial–attentional clustering ensures that each centroid $\mathbf{c}_{j}$ encapsulates both visual and relational coherence.  
The set $\{\mathbf{c}_{j}\}$ forms a preliminary pool of \emph{semantic token candidates} that summarize multi-scale information.

\paragraph{(c) Feature Refinement and Selection.}
To further enhance semantic compactness and eliminate noisy clusters, AdaTA computes composite quality scores for them:
\begin{equation}
s_{j} = 
\eta_{1}\,\mathcal{S}_{\text{size}}(j)
+ \eta_{2}\,\mathcal{S}_{\text{coh}}(j)
- \eta_{3}\,\mathcal{S}_{\text{disp}}(j),
\end{equation}
where 
$\mathcal{S}_{\text{size}}$ measures spatial support,
$\mathcal{S}_{\text{coh}}$ quantifies semantic homogeneity within the cluster, 
and $\mathcal{S}_{\text{disp}}$ penalizes overly scattered distributions.
The top-$K$ clusters with the highest scores are
transformed into \emph{semantic tokens}:
\begin{equation}
\mathbf{S} = \left\{\, \mathbf{c}_{j} \;\middle|\; j \in \operatorname{TopK}(s_j, K) \,\right\}
\end{equation}
After that, the semantic tokens $\mathbf{S}$ are concatenated with the original pixel-level tokens and projected through the multimodal adapter before being integrated with text embeddings in the unified forward path of the LLM.  
Through this adaptive aggregation and refinement process, AdaTA effectively balances global abstraction and local fidelity, enabling the model to reason seamlessly across different granularity.

%% file: sec/4_exp.tex
\section{Experiment}

\newcommand{\dataset}[2]{\mbox{#1\textsubscript{#2}}}

\begin{table*}[t]
\centering
\caption{Comparison of various visual encoders and LLMs across different datasets. Best results are \textbf{bold}.}
\label{tab:comparison_results}
\vspace{-0.5em}
\renewcommand\tabcolsep{4.5pt}
\renewcommand\arraystretch{1.8}

\fontsize{15}{13}

\resizebox{\linewidth}{!}{%
\begin{tabular}{c|c|cc|cc|cc|cc}
\Xhline{1.2pt}
\cellcolor{white}\multirow{2}{*}{\textbf{Encoders}}
& \cellcolor{white}\multirow{2}{*}{\textbf{Models}}
& \multicolumn{2}{c|}{\cellcolor{CadetBlue!20}\textbf{SEED-Bench}}
& \multicolumn{2}{c|}{\cellcolor{CadetBlue!20}\textbf{A-OKVQA}}
& \multicolumn{2}{c|}{\cellcolor{CadetBlue!20}\textbf{Image Caption}}
& \multicolumn{2}{c}{\cellcolor{CadetBlue!20}\textbf{Reasoning}} \\
& & \cellcolor{CadetBlue!20}Recall(\%)$\uparrow$ 
& \cellcolor{CadetBlue!20}GPTscore(\%)$\uparrow$ 
& \cellcolor{CadetBlue!20}Recall(\%)$\uparrow$ 
& \cellcolor{CadetBlue!20}GPTscore(\%)$\uparrow$ 
& \cellcolor{CadetBlue!20}BERTscore(\%)$\uparrow$
& \cellcolor{CadetBlue!20}GPTscore(\%)$\uparrow$
& \cellcolor{CadetBlue!20}BERTscore(\%)$\uparrow$
& \cellcolor{CadetBlue!20}GPTscore(\%)$\uparrow$ \\
\Xhline{1.2pt}

\rowcolor{gray!10}
{\fontsize{14}{16}\selectfont \texttt{CLIP}}
& {\fontsize{14}{16}\selectfont Qwen2.5}
& $50.91_{\color{ForestGreen}{\uparrow 7.89}}$ 
& $24.4_{\color{ForestGreen}{\uparrow 11.4}}$
& $21.79_{\color{ForestGreen}{\uparrow 35.34}}$ 
& $10.3_{\color{ForestGreen}{\uparrow 25.6}}$ 
& $\mathbf{84.35}_{\color{red}{\downarrow 0.04}}$ 
& $21.54_{\color{ForestGreen}{\uparrow 9.74}}$
& $87.79_{\color{ForestGreen}{\uparrow 1.15}}$ 
& $29.36_{\color{ForestGreen}{\uparrow 19.95}}$ \\

\rowcolor{white}
{\fontsize{14}{16}\selectfont \texttt{SigLIP}}
& {\fontsize{14}{16}\selectfont Qwen2.5}
& $46.72_{\color{ForestGreen}{\uparrow 12.08}}$ 
& $23.1_{\color{ForestGreen}{\uparrow 12.7}}$
& $21.89_{\color{ForestGreen}{\uparrow 35.24}}$ 
& $5.1_{\color{ForestGreen}{\uparrow 30.8}}$ 
& $83.92_{\color{ForestGreen}{\uparrow 0.39}}$ 
& $13.59_{\color{ForestGreen}{\uparrow 17.69}}$
& $86.51_{\color{ForestGreen}{\uparrow 2.43}}$ 
& $23.59_{\color{ForestGreen}{\uparrow 25.72}}$ \\

\rowcolor{gray!10}
{\fontsize{14}{16}\selectfont \texttt{DINOv2}}
& {\fontsize{14}{16}\selectfont Qwen2.5}
& $41.40_{\color{ForestGreen}{\uparrow 17.40}}$ 
& $17.7_{\color{ForestGreen}{\uparrow 18.1}}$ 
& $16.67_{\color{ForestGreen}{\uparrow 40.46}}$ 
& $14.1_{\color{ForestGreen}{\uparrow 21.8}}$ 
& $83.40_{\color{ForestGreen}{\uparrow 0.91}}$ 
& $14.36_{\color{ForestGreen}{\uparrow 16.92}}$ 
& $88.15_{\color{ForestGreen}{\uparrow 0.79}}$ 
& $36.67_{\color{ForestGreen}{\uparrow 12.64}}$ \\

\rowcolor{white}
{\fontsize{14}{16}\selectfont \texttt{DINOv3}}
& {\fontsize{14}{16}\selectfont Qwen2.5}
& $55.74_{\color{ForestGreen}{\uparrow 3.06}}$ 
& $15.4_{\color{ForestGreen}{\uparrow 20.4}}$
& $47.43_{\color{ForestGreen}{\uparrow 9.70}}$ 
& $25.6_{\color{ForestGreen}{\uparrow 10.3}}$ 
& $83.90_{\color{ForestGreen}{\uparrow 0.41}}$ 
& $23.97_{\color{ForestGreen}{\uparrow 7.31}}$
& $88.26_{\color{ForestGreen}{\uparrow 0.68}}$ 
& $45.31_{\color{ForestGreen}{\uparrow 4.00}}$ \\

\Xhline{1.2pt}
\rowcolor{yellow!20}
\textbf{Ours}
& {\fontsize{14}{16}\selectfont Qwen2.5}
& $\mathbf{58.80}$ 
& $\mathbf{35.8}$ 
& $\mathbf{57.13}$ 
& $\mathbf{35.9}$ 
& $84.31$ 
& $\mathbf{31.28}$ 
& $\mathbf{88.94}$  
& $\mathbf{49.31}$ \\
\Xhline{1.2pt}

\rowcolor{gray!10}
{\fontsize{14}{16}\selectfont \texttt{CLIP}}
& {\fontsize{14}{16}\selectfont Llama3.2}
& $42.53_{\color{ForestGreen}{\uparrow 6.45}}$ 
& $21.1_{\color{ForestGreen}{\uparrow 32.7}}$
& $25.64_{\color{ForestGreen}{\uparrow 28.97}}$ 
& $25.6_{\color{ForestGreen}{\uparrow 25.7}}$ 
& $82.91_{\color{ForestGreen}{\uparrow 0.91}}$ 
& $7.44_{\color{ForestGreen}{\uparrow 10.64}}$
& $87.01_{\color{ForestGreen}{\uparrow 0.32}}$ 
& $28.97_{\color{ForestGreen}{\uparrow 27.70}}$ \\

\rowcolor{white}
{\fontsize{14}{16}\selectfont \texttt{SigLIP}}
& {\fontsize{14}{16}\selectfont Llama3.2}
& $40.62_{\color{ForestGreen}{\uparrow 8.36}}$ 
& $17.9_{\color{ForestGreen}{\uparrow 35.9}}$
& $17.94_{\color{ForestGreen}{\uparrow 36.67}}$ 
& $6.4_{\color{ForestGreen}{\uparrow 44.9}}$ 
& $\mathbf{84.89}_{\color{red}{\downarrow 1.07}}$ 
& $15.36_{\color{ForestGreen}{\uparrow 2.72}}$
& $\mathbf{88.46}_{\color{red}{\downarrow 1.13}}$ 
& $44.87_{\color{ForestGreen}{\uparrow 11.80}}$ \\

\rowcolor{gray!10}
{\fontsize{14}{16}\selectfont \texttt{DINOv2}}
& {\fontsize{14}{16}\selectfont Llama3.2}
& $40.26_{\color{ForestGreen}{\uparrow 8.72}}$ 
& $23.1_{\color{ForestGreen}{\uparrow 30.7}}$ 
& $35.90_{\color{ForestGreen}{\uparrow 18.71}}$ 
& $17.9_{\color{ForestGreen}{\uparrow 33.4}}$ 
& $82.59_{\color{ForestGreen}{\uparrow 1.23}}$ 
& $5.54_{\color{ForestGreen}{\uparrow 12.54}}$ 
& $86.55_{\color{ForestGreen}{\uparrow 0.78}}$ 
& $19.49_{\color{ForestGreen}{\uparrow 37.18}}$ \\

\rowcolor{white}
{\fontsize{14}{16}\selectfont \texttt{DINOv3}}
& {\fontsize{14}{16}\selectfont Llama3.2}
& $47.23_{\color{ForestGreen}{\uparrow 1.75}}$ 
& $30.3_{\color{ForestGreen}{\uparrow 23.5}}$
& $32.05_{\color{ForestGreen}{\uparrow 22.56}}$ 
& $7.7_{\color{ForestGreen}{\uparrow 43.6}}$ 
& $82.98_{\color{ForestGreen}{\uparrow 0.84}}$ 
& $8.12_{\color{ForestGreen}{\uparrow 9.96}}$
& $87.26_{\color{ForestGreen}{\uparrow 0.07}}$ 
& $34.87_{\color{ForestGreen}{\uparrow 21.80}}$ \\

\Xhline{1.2pt}
\rowcolor{yellow!20}
\textbf{Ours}
& {\fontsize{14}{16}\selectfont Llama3.2}
& $\mathbf{48.98}$ 
& $\mathbf{53.8}$ 
& $\mathbf{54.61}$ 
& $\mathbf{51.3}$ 
& $83.82$ 
& $\mathbf{18.08}$ 
& $87.33$  
& $\mathbf{56.67}$ \\
\Xhline{1.2pt}

\end{tabular}%
}
\end{table*}

\subsection{Experiments Setup}
\paragraph{Datasets \& Metrics.}
We evaluate \texttt{Granulon} across \textbf{5 benchmarks} to investigate how different visual encoders affect MLLM performance under varying task demands: \ding{182} \textbf{VQA}: SEED-Bench \cite{li2023seed} and A-OKVQA \cite{schwenk2022okvqa}; \ding{183} \textbf{Image Caption}: CC12M$\_$and$\_$Imagenet21K$\_$Recap \cite{changpinyo2021conceptual}; \ding{184} \textbf{Reasoning}: FLUX-Reason \cite{fang2025flux}; \ding{185} \textbf{Medical Domain}: SurgVLM \cite{zeng2025surgvlm}. We employ complementary metrics to assess multimodal understanding and generation: GPT-4o serves as an automatic judge \cite{li2025generation, li2024llms} for \textit{semantic accuracy}, \textit{hallucination rate} and \textit{granularity score};
Recall is reported for VQA to reflect factual correctness, and BERTscore is computed to evaluate semantic fidelity and linguistic consistency. 
More details are provided in the Appendix \ref{app:a}.

\begin{table}[t] 
\caption{Comparison of various visual encoders across different medical tasks. Best results are \textbf{bold}.}
\label{tab:surgical}
\vspace{-0.5em}
\renewcommand\tabcolsep{3.5pt} 
\renewcommand\arraystretch{1.6} %
\fontsize{9}{11}\selectfont %
\resizebox{\linewidth}{!}{%
\begin{tabular}{c|cc|cc}
\Xhline{1.2pt}
\cellcolor{white}\multirow{2}{*}{Encoders}
& \multicolumn{2}{c|}{\cellcolor{CadetBlue!20}\textbf{Phase Recognition}}
& \multicolumn{2}{c}{\cellcolor{CadetBlue!20}\textbf{Instrument Recognition}} \\
& \cellcolor{CadetBlue!20}BERTscore(\%)$\uparrow$
& \cellcolor{CadetBlue!20}Recall(\%)$\uparrow$
& \cellcolor{CadetBlue!20}BERTscore(\%)$\uparrow$
& \cellcolor{CadetBlue!20}Recall(\%)$\uparrow$ \\
\Xhline{1.2pt}

\rowcolor{gray!10}
\texttt{CLIP}
& $91.64_{\color{ForestGreen}{\uparrow 5.68}}$ 
& $46.15_{\color{ForestGreen}{\uparrow 30.77}}$
& $94.44_{\color{ForestGreen}{\uparrow 3.51}}$ 
& $46.92_{\color{ForestGreen}{\uparrow 29.15}}$ \\

\rowcolor{white}
\texttt{SigLIP}
& $91.98_{\color{ForestGreen}{\uparrow 5.34}}$ 
& $48.72_{\color{ForestGreen}{\uparrow 28.20}}$
& $95.71_{\color{ForestGreen}{\uparrow 2.04}}$ 
& $\mathbf{78.20}_{\color{red}{\downarrow 2.13}}$ \\

\rowcolor{gray!10}
\texttt{DINOv2}
& $91.67_{\color{ForestGreen}{\uparrow 5.65}}$ 
& $33.33_{\color{ForestGreen}{\uparrow 43.59}}$ 
& $95.93_{\color{ForestGreen}{\uparrow 2.02}}$ 
& $55.21_{\color{ForestGreen}{\uparrow 20.86}}$ \\

\rowcolor{white}
\texttt{DINOv3}
& $94.71_{\color{ForestGreen}{\uparrow 2.61}}$ 
& $64.10_{\color{ForestGreen}{\uparrow 12.82}}$
& $97.41_{\color{ForestGreen}{\uparrow 0.54}}$ 
& $68.89_{\color{ForestGreen}{\uparrow 7.18}}$ \\

\Xhline{1.2pt}
\rowcolor{yellow!20}
\textbf{Ours}
& $\mathbf{97.32}$ 
& $\mathbf{76.92}$ 
& $\mathbf{97.95}$  
& $76.07$ \\
\Xhline{1.2pt}

\end{tabular}%
}
\vspace{-1.0 em}
\end{table}

\vspace{-1.0 em}
\paragraph{Implementation Details.}%
All experiments are conducted under the LLaVA framework, where the visual encoder is replaced while keeping the remaining architecture, datasets, model size, and optimization hyperparameters \textbf{identical}. 
We integrate several visual encoders with two language backbones: \textit{Qwen-2.5-Instruct-1.5B} and \textit{Llama-3.2-3B}. 
Training is performed on 8$\times$H200 GPUs for 2 epochs using a batch size of 128 and a learning rate of $2\times10^{-5}$. 
Further details are provided in the Appendix \ref{app:a}.

\subsection{Main Results}

\paragraph{High-performing.}
As summarized in Table~\ref{tab:comparison_results}, \textbf{\texttt{Granu-}} \textbf{\texttt{lon}} consistently surpasses all \texttt{CLIP}- and \texttt{DINO}-based MLLMs under identical training and evaluation settings.
Using Qwen2.5 as the LLM backbone, \texttt{Granulon} achieves \(58.8\%\) and \(57.13\%\) Recall on SEED-Bench and A-OKVQA, outperforming the \texttt{CLIP} baseline by \(+7.89\%\) and \(+35.34\%\), respectively.
On image captioning, it reaches a GPTscore of \(31.28\%\), exceeding \texttt{SigLIP} and \texttt{DINOv2} by \(+17.69\%\) and \(+7.31\%\) points.
These results demonstrate that adaptive granularity aggregation effectively enhances multimodal understanding in a unified forward pass.

\begin{figure}[h]
\centering
\includegraphics[width=1.0\linewidth]{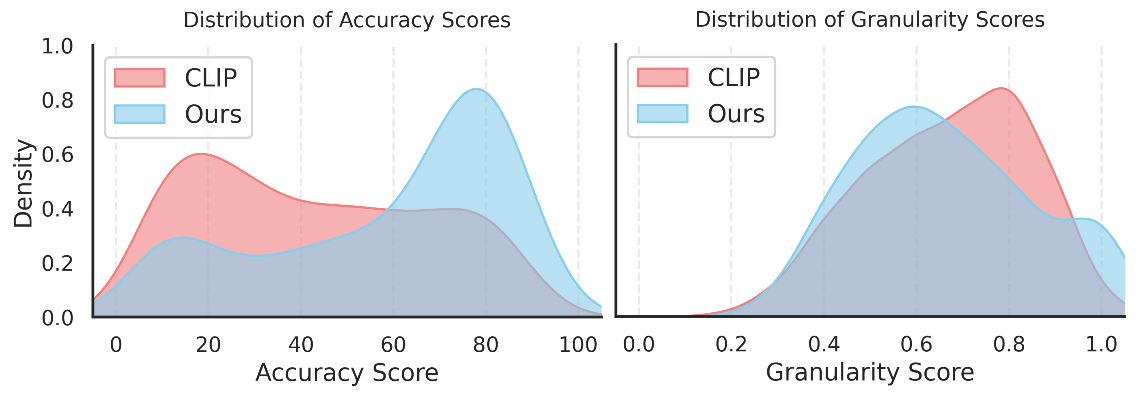}
\caption{Distribution of accuracy and granularity obtained from reasoning outputs. We summarize results across 120 samples, where the granularity and accuracy scores of each sentence are collected. Lower granularity score means finer-grained outputs.}
\label{fig:density}
\vspace{-0.5 em}
\end{figure}
\paragraph{Fine-grained Reasoning.}
\texttt{Granulon} exhibits clear advantages in reasoning benchmarks that demand contextual and compositional inference.
On FLUX-Reason, the model achieves a GPTscore of \(56.67\%\) with Llama3, outperforming \texttt{DINOv2} (\(19.49\%\)) and \texttt{CLIP} (\(28.97\%\)) by \(37.18\%\!\uparrow\) and \(27.70\%\!\uparrow\), respectively. 
As illustrated in Figure~\ref{fig:density}, the distribution of reasoning accuracy and granularity in \texttt{Granulon} is more concentrated in the high-accuracy and fine-grained region, whereas \texttt{CLIP} tends to operate at coarser levels with lower overall accuracy.
These indicate that \texttt{Granulon} delivers stronger fine-grained comprehension and richer hierarchical granularity, capable of capturing fine local details and improving multi-step reasoning while maintaining global semantic consistency.

\vspace{-1 em}
\paragraph{Generalization to medical domain.}
To further assess the cross-domain capability of \texttt{Granulon}, we evaluate it on fine-grained recognition tasks in medical domain that require distinguishing subtle visual details. 
As shown in Table~\ref{tab:surgical}, \texttt{Granulon} achieves superior results in both phase recognition and instrument recognition, reaching \(97.32\%\) and \(97.95\%\) BERTscore. 
Compared to \texttt{CLIP} and \texttt{DINOv3}, \texttt{Granulon} also produces notable gains of \(+30.77\%\) and \(+12.82\%\) in Recall, respectively. 
These results demonstrate that adaptive granularity effectively preserves discriminative details while maintaining robust global understanding, enabling strong generalization in domain-specific scenarios.

\begin{figure}[t]
\centering
\includegraphics[width=1.0\linewidth]{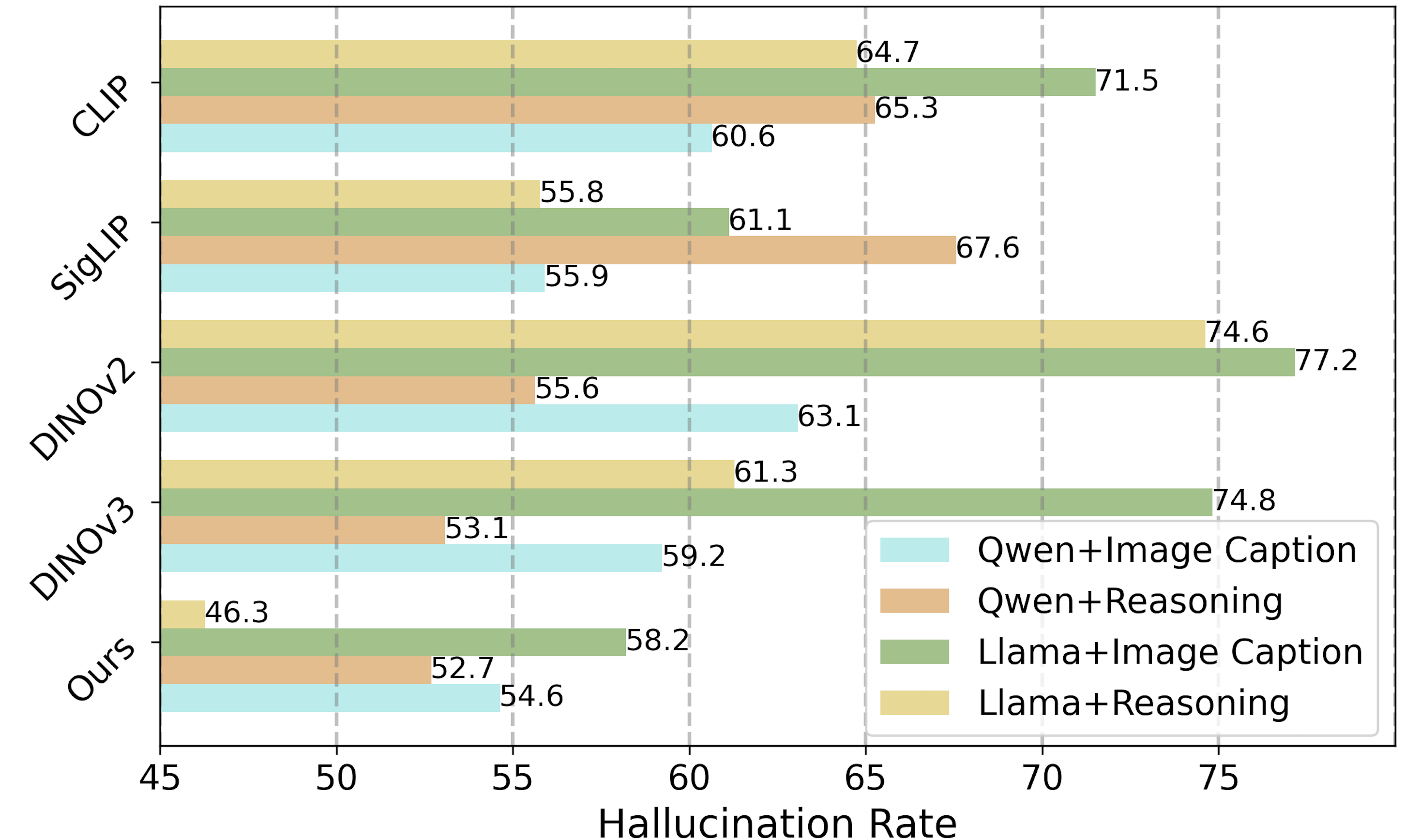}
\caption{Hallucination rate of various visual encoder.}
\label{fig:hallucination}
\vspace{-1.2 em}
\end{figure}

\subsection{Robustness and Trustworthiness}
\paragraph{Hallucination Resistance.}
As shown in Figure~\ref{fig:hallucination}, \texttt{Granulon} achieves the lowest hallucination rate among all visual encoders on both Qwen2.5 and Llama3.
On Image Caption, \texttt{Granulon} reduces hallucination by \(6.0\%\) and \(4.6\%\) relative to \texttt{CLIP} and \texttt{DINOv3} with Qwen.
Additionally, in the reasoning tasks with Llama3, the rate drops from \(61.3\%\) (\texttt{DINOv3}) to \(46.3\%\) (\texttt{Granulon}), indicating a relative reduction of \(46\%\).
Such improvements highlight that fine-to-coarse granularity alignment helps the LLM balance detail preservation with semantic coherence, effectively reducing hallucination propagation.

\begin{figure}[t]
\centering
\includegraphics[width=1.0\linewidth]{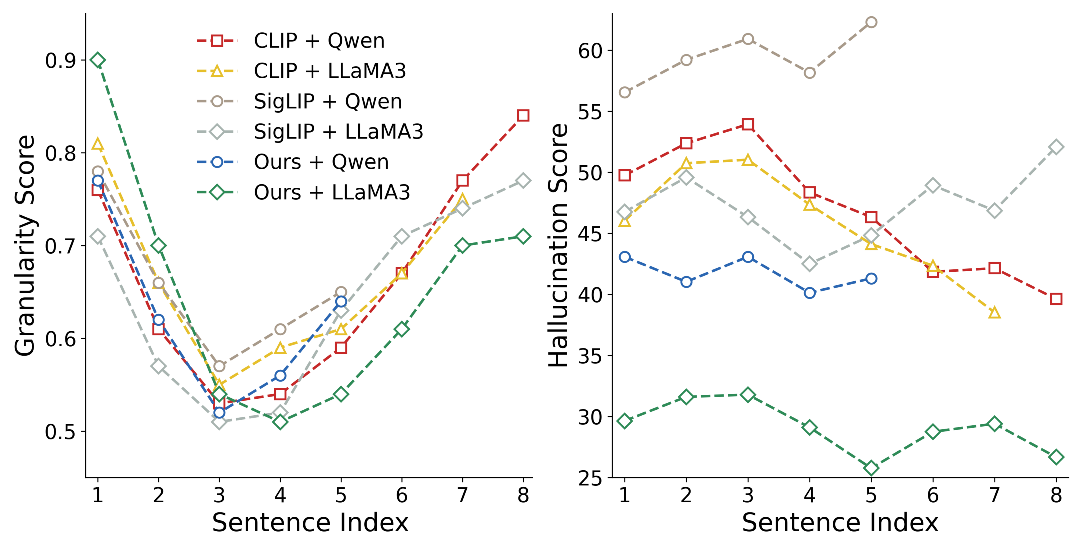}

\caption{Comparison of sentence-level granularity and hallucination scores across different visual encoders. Sentence index means the n-th sentence generated by LLMs.}
\label{fig:sentence}
\vspace{-1.8 em}
\end{figure}

\vspace{-1.3 em}
\paragraph{Granularity \& Hallucination Analysis.}
To further investigate the relationship between visual granularity and hallucination, we analyze 240 cases and compute both scores for each sentence in Figure~\ref{fig:sentence}.
We observe that \ding{182} \texttt{CLIP} baselines exhibit unstable semantic consistency, losing fine details in fine-grained stage while slightly recovering in coarse sentences. \ding{183} \texttt{SigLIP} maintains relatively stable granularity but achieves lower overall scores, suggesting weaker alignment between visual granularity and semantic consistency.
In contrast, \ding{184} \texttt{Granulon} achieves consistently low hallucination scores, indicating that it captures multi-level visual semantics more stably and preserves detail without sacrificing factual grounding.

\begin{figure}[t]
\centering
\includegraphics[width=1.0\linewidth]{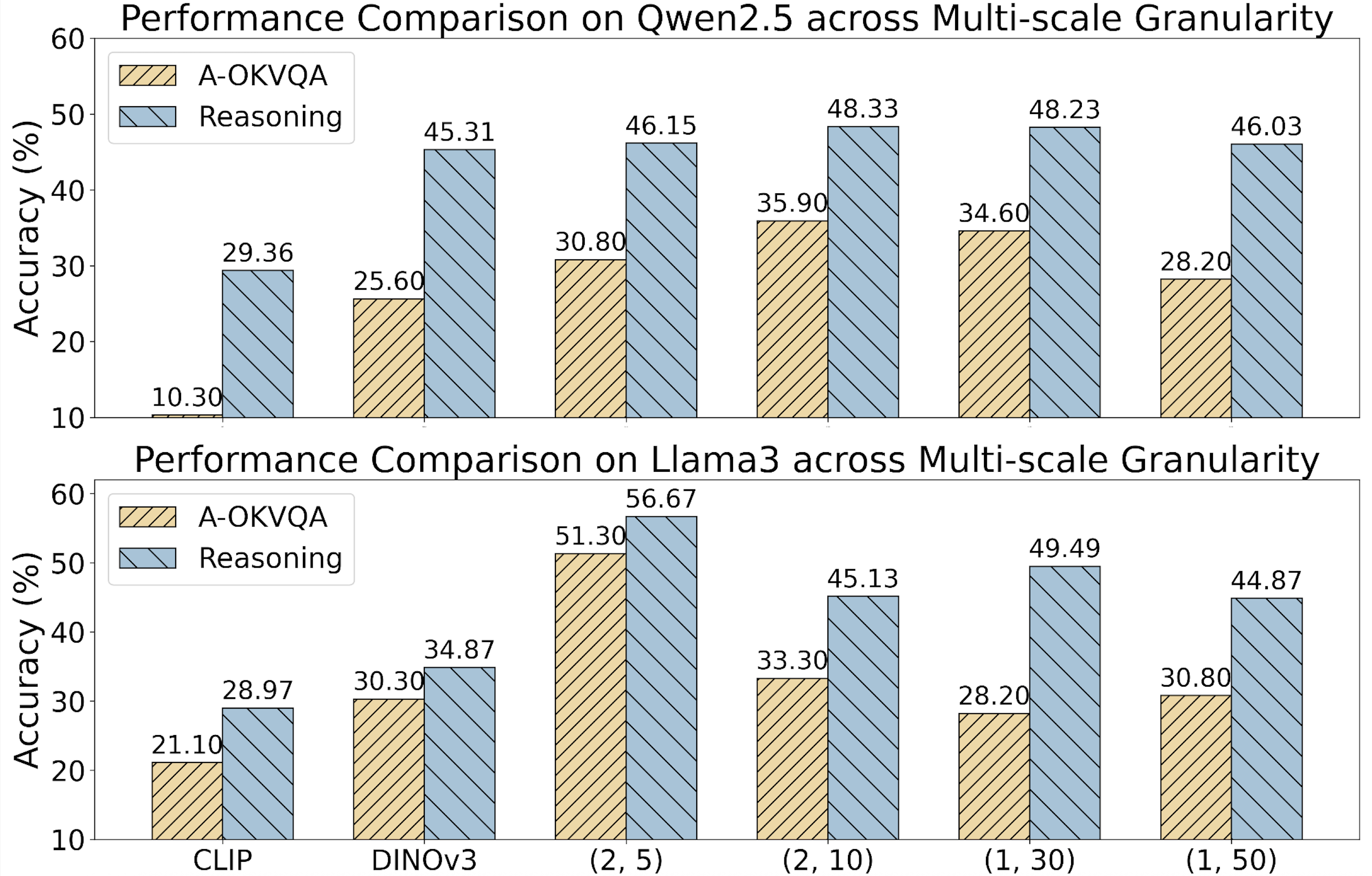}
\caption{Performance Comparison of multi-scale granularity.}
\label{fig:ablation}
\vspace{-1.5 em}
\end{figure}

\subsection{Ablation Studies}
\paragraph{Effect of Semantic Token Granularity.}
We examine the impact of semantic token granularity within the AdaTA module under A-OKVQA and FLUX-Reason in Figure~\ref{fig:ablation}.
The x-axis denotes the pooling kernel size and corresponding cluster number for semantic token aggregation.
In both tasks, all granularity-aware variants outperform the plain \texttt{DINOv3} baseline, confirming the effectiveness of adaptive abstraction.
In the coarse A-OKVQA setting, coarser configurations with fewer clusters yield superior performance. For instance, Accuracy (GPTscore) improves by \(\sim\!20\%\) with only 5 clusters, while excessive fine-grained clustering with 50 clusters has merely a \(0.50\%\!\uparrow\) gain due to over-fragmented semantics.
Conversely, reasoning tasks benefit from finer granularity, with GPTScore increasing as the cluster count grows (\(\sim35\%\!\rightarrow\sim45\%\)).
These results indicate that the optimal level of granularity is task-dependent: coarse abstraction benefits global understanding, while fine abstraction supports detailed reasoning.

\begin{figure}[t]
\centering
\includegraphics[width=1.0\linewidth]{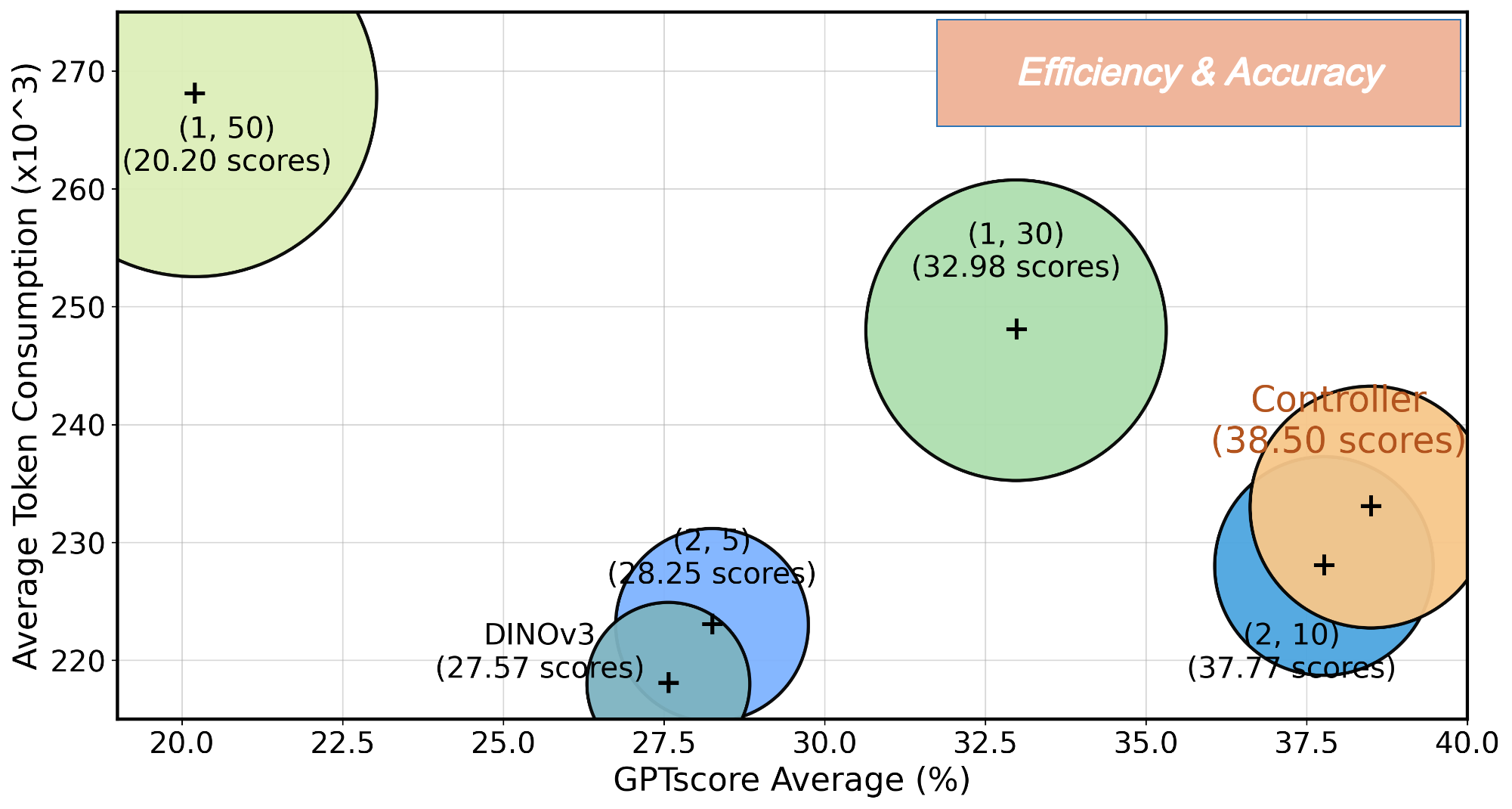}
\vspace{-2 em}
\caption{Visualization of the performance metrics and input
token consumption of different scale.}
\label{fig:token}
\vspace{-0.5 em}
\end{figure}

\vspace{-1.3 em}
\paragraph{Impact of the Granularity Controller (Efficiency \& Accuracy).}
We further analyze the effect of the text-conditioned Controller by comparing models with and without adaptive granularity selection (Figure~\ref{fig:token}).
The fixed-token variant shows noticeable degradation, with average GPTscore dropping compared to the adaptive version.
Additionally, although the adaptive Controller introduces around \(10\%\) higher token consumption, \texttt{Granulon} achieves \(38.50\%\), representing a substantial \(+39.7\%\) improvement compared to the vanilla \texttt{DINOv3}.
This indicates that the key factor behind the improvement is not token quantity, but \emph{text-adaptive granularity selection}:  
by adjusting the visual abstraction level according to the question semantics, \texttt{Granulon} achieves both higher precision and better computational efficiency.

\subsection{Qualitative and Analytical Insights}

\begin{figure}[t]
\centering
\includegraphics[width=1.0\linewidth]{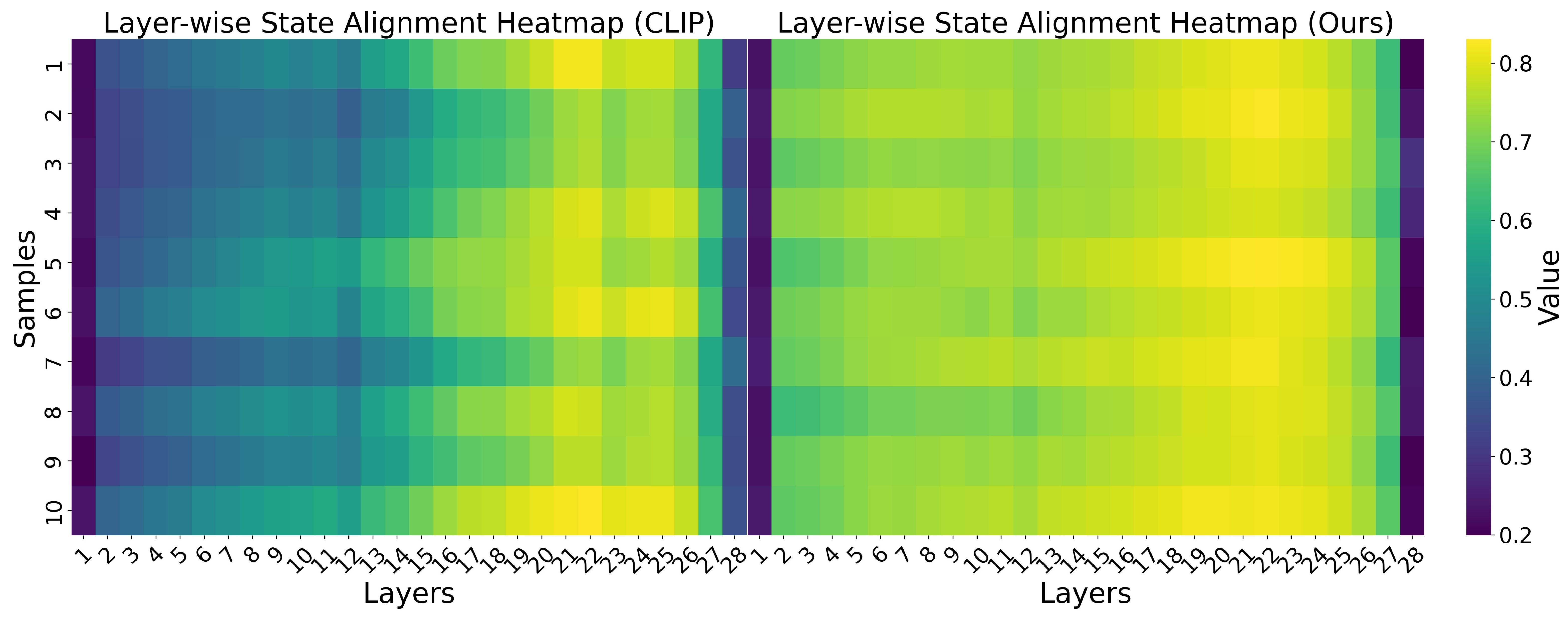}
\caption{
    We first only input the reference texts into the LLM to obtain the hidden states at each layer, 
    and then compute the cosine similarity between the reference state and the state of each corresponding layer for both \texttt{CLIP} baseline and \texttt{Granulon}.
    }
\label{fig:heatmap}
\vspace{-1.5 em}
\end{figure}

\paragraph{Layer-wise Alignment Analysis.}
To isolate the representational dynamics of the visual encoders from the specific routing behaviors of the LLMs, we examine the layer-wise state alignment under Llama3. As illustrated in Figure \ref{fig:heatmap}, the \texttt{CLIP}-based model establishes a solid initial semantic foundation, with cosine similarity scores stabilizing \(\sim0.60\). While highly effective for global semantic grounding, its fixed-resolution nature restricts further alignment progression, as the LLM increasingly demands fine-grained visual evidence during deep compositional reasoning. Conversely, \texttt{Granulon} exhibits a continuous strengthening of cross-layer alignment, reaching a similarity ceiling of \(\sim0.80\). By dynamically integrating pixel-level fidelity with adaptive semantic abstraction, \texttt{Granulon} provides a multi-scale representational scaffold that seamlessly supports the hierarchical computation of LLM. This consistent divergence, identically observed across alternative backbones in Figures~\ref{fig:clip_llama}–\ref{fig:ours_qwen}, confirms that the superior semantic retention \texttt{Granulon} stems intrinsically from its adaptive granularity rather than language model biases.

\vspace{-1.3 em}
\paragraph{Qualitative Case Study.}
To illustrate the difference in reasoning behavior between \texttt{CLIP} and \texttt{Granulon}, we present a representative case study below. Text in \textcolor{green!50!black}{green} marks correctly grounded content, while \textcolor{red!70!black}{red} denotes hallucinated descriptions.
As observed, \ding{110} \texttt{CLIP} tends to produce semantically related yet unsupported details in the latter part of its response. This phenomenon arises from its semantic-level visual representation, which encourages LLMs to associate visual cues with contextually irrelevant concepts. 
In contrast, \ding{110} \texttt{Granulon} maintains consistent and faithful reasoning throughout the output. We attribute this robustness to its pixel-level feature encoding, which preserves fine-grained visual cues and prevents LLMs from over-activating internal semantic priors, thereby mitigating hallucination and maintaining semantic fidelity. 

\begin{tcolorbox}[colback=gray!10!white, colframe=gray!80!black, title=Case Study: Reasoning Answer]

\small\itshape   %
\setlength{\intextsep}{12pt}
\setlength{\columnsep}{10pt}
\begin{wrapfigure}{l}{0.35\linewidth} %
    \vspace{-1em}
    \includegraphics[width=\linewidth]{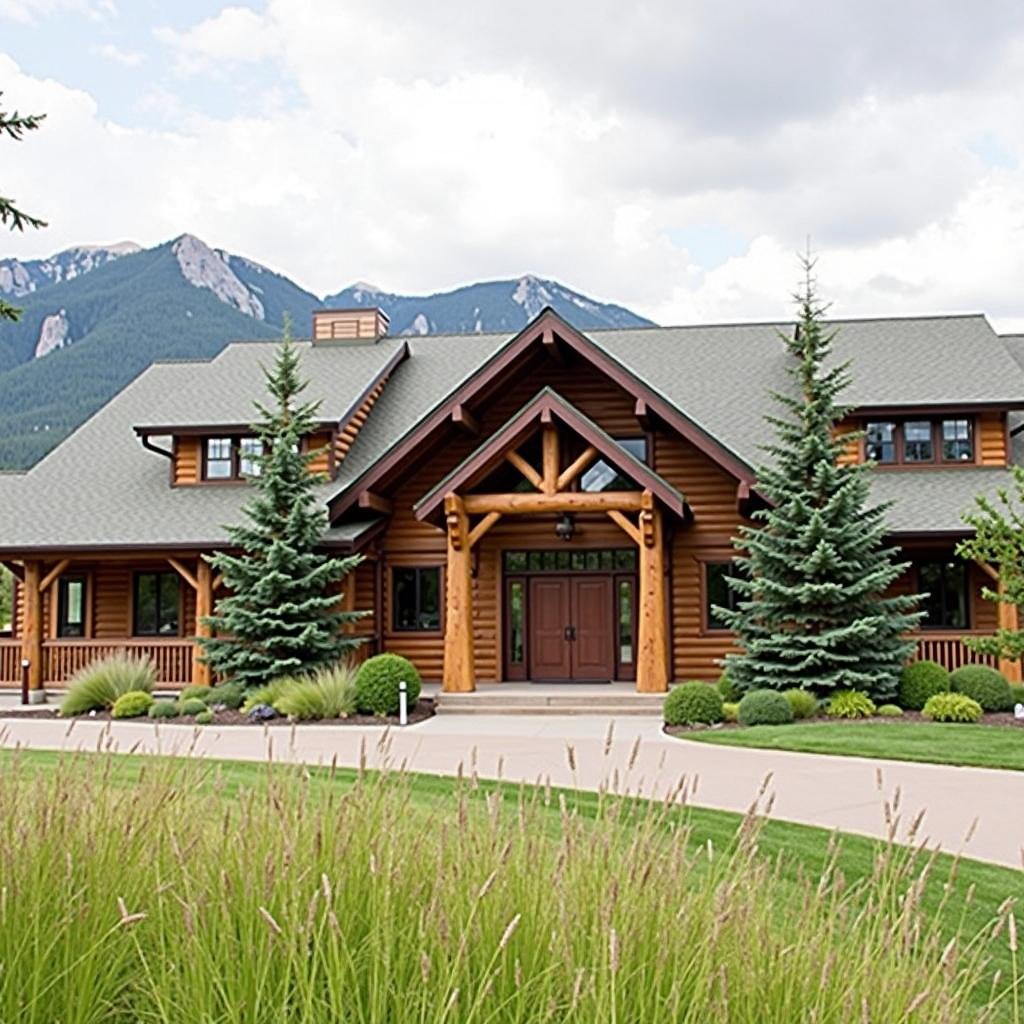}
    \vspace{-2em}
\end{wrapfigure}

\textbf{\texttt{CLIP}:} \textellipsis The front door itself is a focal point, featuring \textcolor{green!50!black}{a large, arched window}, adding a touch of elegance and sophistication. \textcolor{red!70!black}{The interior} is designed with \textcolor{red!70!black}{a blend of modern and rustic elements}, showcasing a combination of \textcolor{red!70!black}{hardwood floors}, \textcolor{red!70!black}{polished stone}, and \textcolor{red!70!black}{natural wood accents}, complemented by a spacious \textcolor{red!70!black}{living room, a dining area, and a gourmet kitchen} with stainless steel appliances. The property also features \textcolor{red!70!black}{a large deck with a pergola overhead}, perfect for outdoor entertaining \textellipsis

\textbf{OURS:} 
Set against a backdrop of \textcolor{green!50!black}{pristine snow-covered mountains}, a \textcolor{green!50!black}{rustic wooden lodge} stands prominently, its distinctive design and warm tones evoking a sense of comfort and tranquility. The lodge, with its \textcolor{green!50!black}{peaked roof} and \textcolor{green!50!black}{overhanging eaves}, features a \textcolor{green!50!black}{central chimney} and \textcolor{green!50!black}{a pair of windows} on either side, adding to its inviting atmosphere. The exterior is adorned with a mix of natural elements, including a row of \textcolor{green!50!black}{neatly trimmed evergreen trees} and a cluster of colorful flowering \textcolor{green!50!black}{shrubs}, creating a harmonious blend of nature and craftsmanship. The surrounding landscape is dotted with small, well-maintained \textcolor{green!50!black}{gardens} and \textcolor{green!50!black}{lawns}, further emphasizing the lodge's peaceful setting.

\end{tcolorbox}

%% file: sec/X_suppl.tex
\section{More Implementation Details}
\label{app:a}

\paragraph{Dataset Processing.}

For general-purpose multimodal reasoning, we use a 200K subset of the FLUX-Reason corpus as our primary training data. This subset preserves the diversity of the original dataset, spanning multi-step reasoning, compositional inference, and fine-grained visual understanding. We additionally hold out another 1K FLUX-Reason examples as an internal validation split for consistent in-domain evaluation under \textbf{identical} training conditions.
To assess zero-shot generalization beyond the training distribution, we rely exclusively on models trained on FLUX-Reason. We construct compact evaluation suites by uniformly sampling 1K examples each from SEED-Bench, A-OKVQA, and CC12M$\_$and$\_$Imagenet21K$\_$Recap. These benchmarks span factual VQA, compositional reasoning, and conceptual caption generation, enabling a comprehensive assessment of cross-domain robustness. For the medical domain, we use a 50K subset of SurgVLM as the training corpus. We also reserve an additional 1K samples as a test split for evaluating medical visual understanding and reasoning.

\paragraph{Training Procedure.}
All models are trained under the LLaVA framework, where different visual encoders are seamlessly integrated with a unified multimodal processing pipeline. All visual encoders used have a parameter scale of approximately 0.4B, ensuring a fair comparison across architectures. To ensure a strictly controlled comparison across encoders, we freeze both the visual backbone and the language model backbone throughout training. Only the multimodal projector, which aligns visual features with the LLM token space, is fully optimized. This setup allows us to isolate and analyze the effect of visual encoder representations while holding all other components constant.

The models are trained on the FLUX-Reason subset for general benchmarks and on the SurgVLM subset for medical benchmarks, with full-parameter optimization applied to the multimodal projector. We employ DeepSpeed ZeRO-2 for memory-optimized distributed training and adopt a two-epoch training schedule with a cosine learning-rate decay. The batch configuration combines a per-device batch size of 8 with gradient accumulation, effectively yielding a global batch size suited for stable optimization. Mixed-precision settings (\texttt{bf16} with \texttt{tf32} matrix acceleration) are used to improve computational efficiency without compromising numerical stability.

The optimization follows an AdamW variant with a learning rate of $1\times10^{-4}$, weight decay set to zero, and a warm-up ratio of 0.05. We disable intermediate evaluations to maximize throughput and checkpoint the model at fixed training intervals. Notably, across all visual encoders, the training curves exhibit highly similar convergence behaviors; the final loss values converge to comparable levels, indicating that loss magnitude alone is insufficient for distinguishing model quality. Instead, downstream reasoning and zero-shot evaluations provide more reliable measurements of visual encoder effectiveness.

\paragraph{Metrics.}
We adopt a two-stage evaluation strategy to assess the performance of different visual encoders on the FLUX-Reason test set and the zero-shot benchmarks. We first compute BERTscore as a baseline similarity metric. However, the scores across models differ only marginally, despite clear qualitative differences in the generated outputs. This highlights a fundamental limitation of embedding-based similarity metrics: their inability to capture correctness, factual grounding, or the completeness of multimodal reasoning.

To obtain more faithful and discriminative evaluations, we employ GPT-4o as an automatic judge \cite{zheng2023judging, li2024llms, li2025generation, tan2024judgebench}. Through extensive manual verification, we find that GPT-4o's assessments align significantly better with human judgment than BERTscore. Importantly, for the FLUX-Reason benchmark, we do \emph{not} compare model outputs against the provided textual ground truth. Although the reference annotations are rich, they inevitably fail to exhaustively enumerate all objects, attributes, and relationships present in an image. In practice, we observe that \texttt{Granulon} frequently produces correct and well-grounded descriptions that extend beyond the reference annotations. As a result, we evaluate responses by directly comparing them against the image content itself, with GPT-4o determining whether the reasoning chain, visual grounding, and final conclusions faithfully reflect what is depicted.
For VQA benchmarks, GPT-4o judges the factual correctness of predicted answers beyond surface-level lexical matches. For image captioning tasks, GPT-4o evaluates semantic coverage, visual relevance, and the ability to highlight salient elements of the scene.

\paragraph{Experimental Configurations for Visual Encoders.}
To ensure a rigorous and fair comparison, we standardize the parameter capacity across all selected visual encoders to a comparable scale of approximately 0.3B to 0.4B parameters. Specifically, for the \texttt{CLIP} baseline, we utilize the \texttt{CLIP} ViT-L/14@336 model, which operates at a higher input resolution of 336$\times$336 pixels. For the DINO-family baselines, we employ \texttt{DINOv3} ViT-L/16 and \texttt{DINOv2} ViT-L/14, both of which provide robust region-level feature representations at a similar model capacity. Similarly, for the \texttt{SigLIP} baseline, we select the \texttt{SigLIP} ViT-L/16 variant. Consistent with our overall methodology, all visual backbones and the downstream language model are strictly frozen during the training phase. Only the multimodal projector is fully optimized to align the distinct global or fine-grained features extracted by each respective encoder with the textual embedding space of the language model.

\section{Evaluation Prompts}
To quantitatively and objectively evaluate the multimodal capabilities of the models, we employed GPT-4o as an unbiased judge. The evaluation is conducted using the following prompt, which explicitly instructs the model to assess both the semantic accuracy and the hallucination rate based on the provided image and the generated prediction. 

\vspace{1em}
\begin{tcolorbox}[
    enhanced,
    colback=promptbg,
    colframe=promptframe,
    coltitle=white,
    fonttitle=\bfseries\large,
    title=GPT-4o Evaluator,
    arc=2mm,
    boxrule=1.5pt,
    left=10pt, right=10pt, top=10pt, bottom=10pt
]
You are an objective evaluator for AI image descriptions.
Evaluate the following MODEL OUTPUT against the PROVIDED IMAGE,
and assign TWO INDEPENDENT scores (0-100 Accuracy, 0-100 Hallucination).

\vspace{0.8em}
1. {ACCURACY SCORE (0-100): how well the text matches the actual image content.\\
2. HALLUCINATION SCORE (0-100): how much content is NOT present in the image.

\vspace{0.8em}
MODEL OUTPUT: \{pred\}

\vspace{0.8em}
Return ONLY a valid JSON object:\\
\verb|{"accuracy_score": int, | \\
\verb|"hallucination_score": int}|
}
\end{tcolorbox}

\section{More Layer-wise Alignment Analysis}
\begin{figure}[h]
\centering
\includegraphics[width=1.0\linewidth]{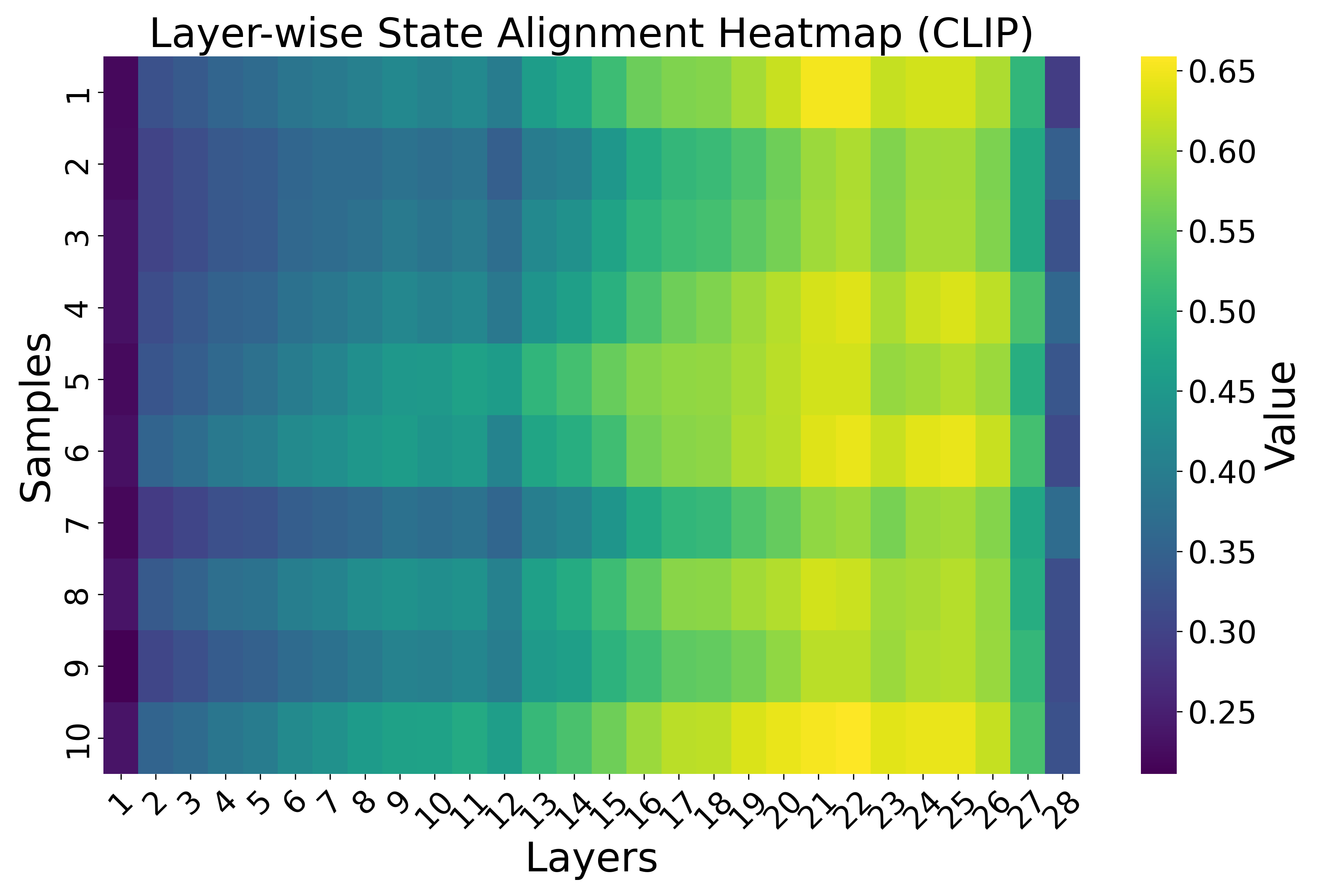}
\caption{Cosine similarity between the reference state and the state of each corresponding layer for \texttt{CLIP}+Llama3.}
\label{fig:clip_llama}
\vspace{1.2 em}
\end{figure}

\begin{figure}[h]
\centering
\includegraphics[width=1.0\linewidth]{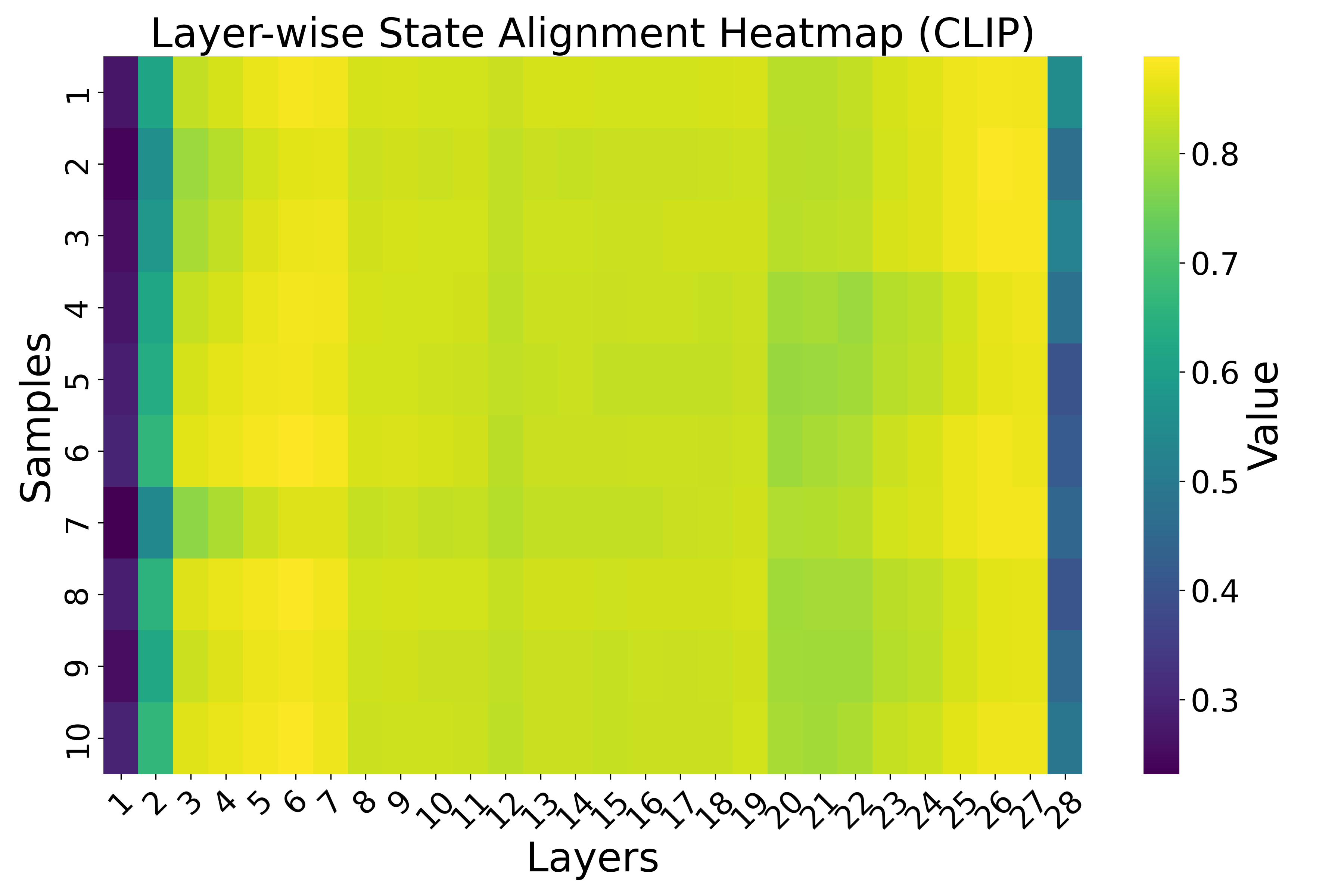}
\caption{Cosine similarity between the reference state and the state of each corresponding layer for \texttt{CLIP}+Qwen2.5.}
\label{fig:clip_qwen}
\vspace{1.2 em}
\end{figure}

\begin{figure}[h]
\centering
\includegraphics[width=1.0\linewidth]{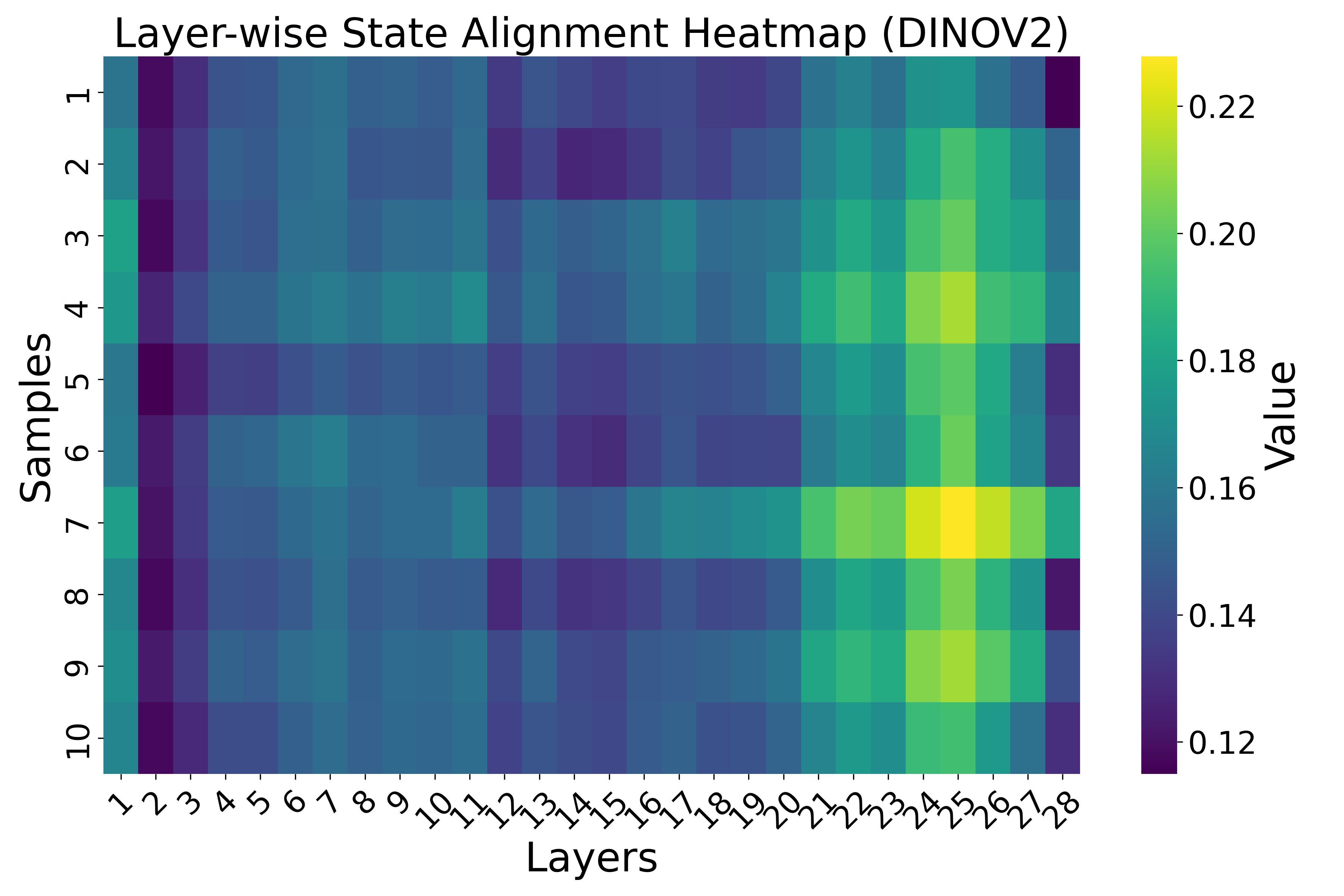}
\caption{Cosine similarity between the reference state and the state of each corresponding layer for \texttt{DINOV2}+Llama3.}
\label{fig:dionv2_llama}
\vspace{1.2 em}
\end{figure}

\begin{figure}[h]
\centering
\includegraphics[width=1.0\linewidth]{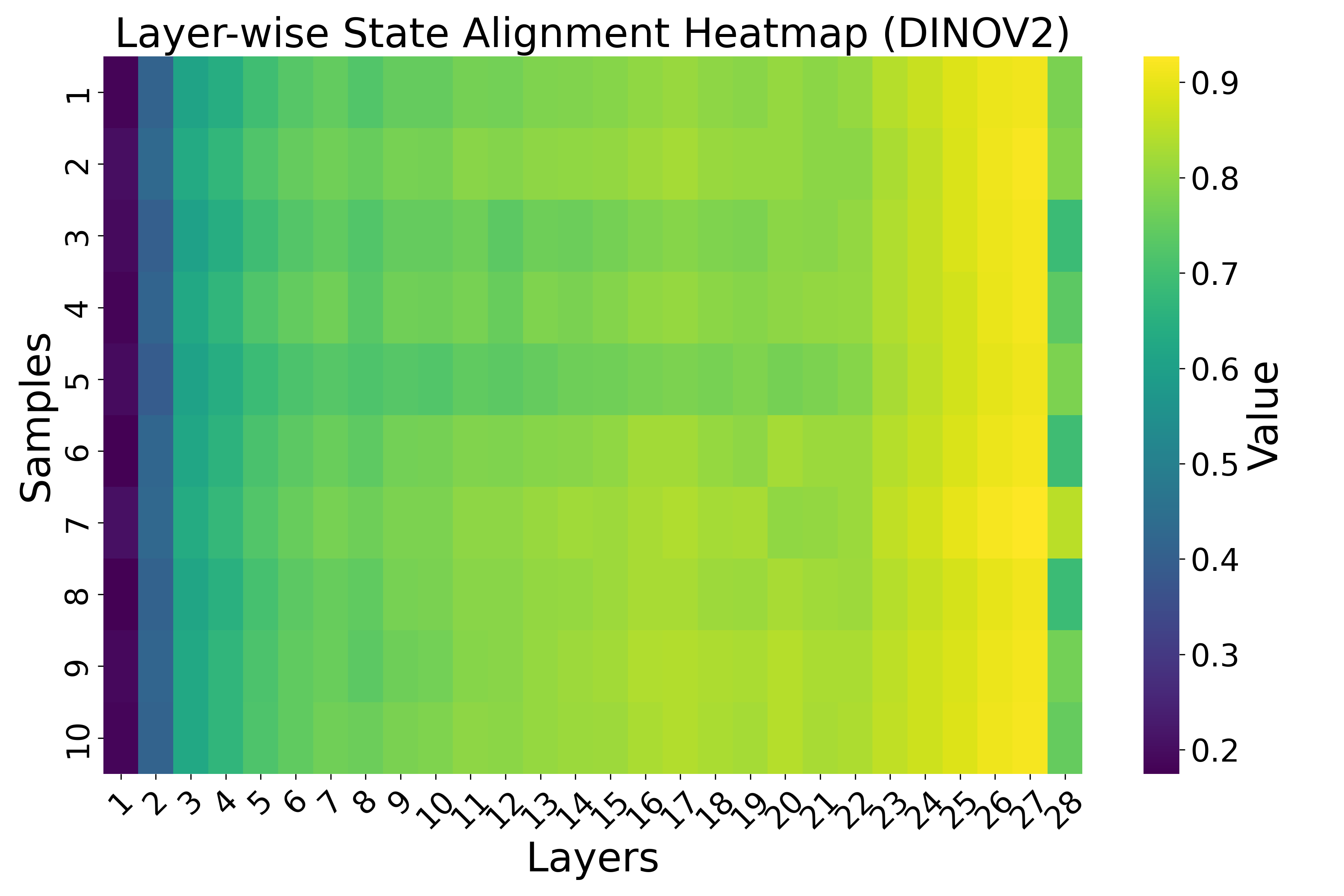}
\caption{Cosine similarity between the reference state and the state of each corresponding layer for \texttt{DINOV2}+Qwen2.5.}
\label{fig:dinov2_qwen}
\end{figure}

\begin{figure}[h]
\centering
\includegraphics[width=1.0\linewidth]{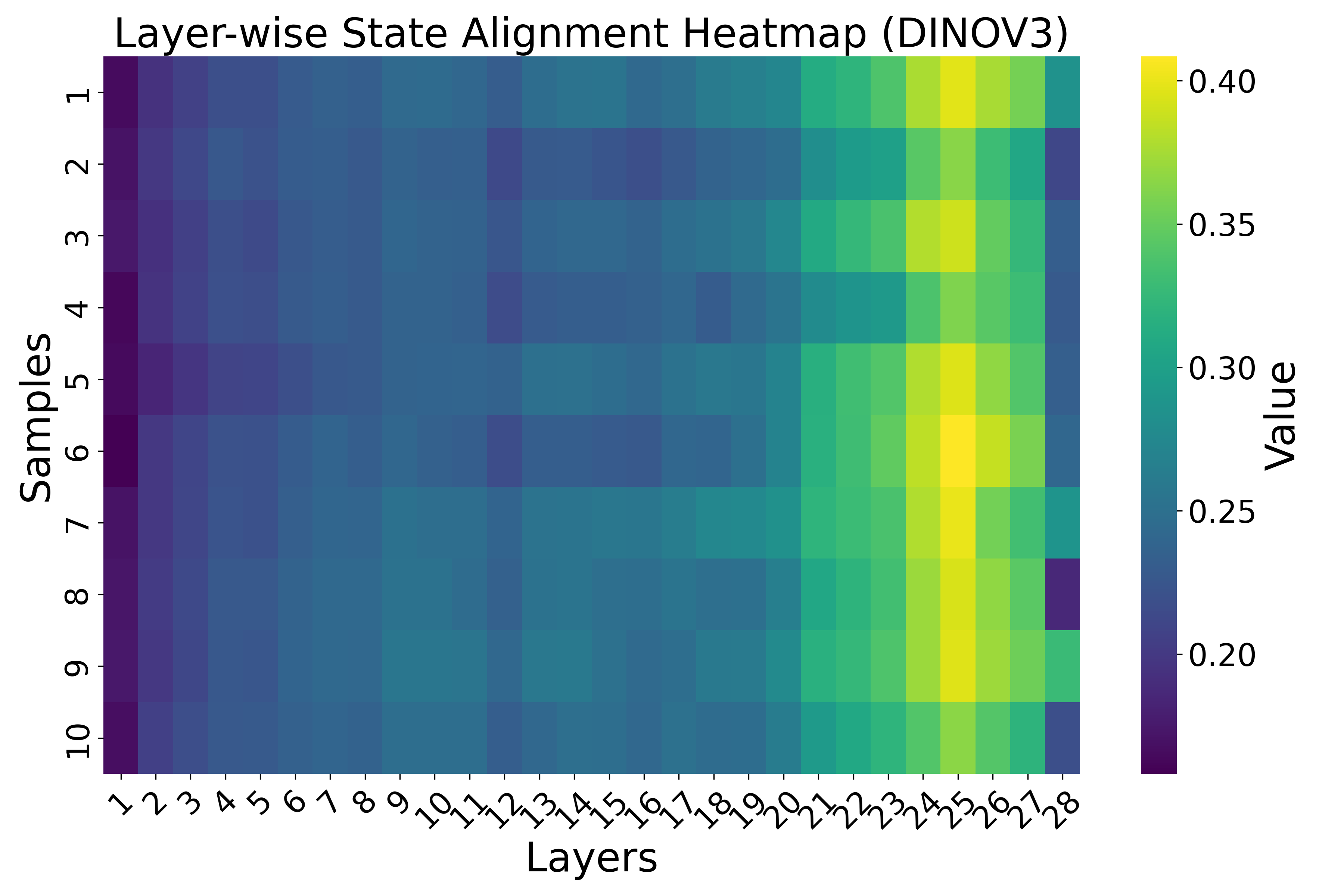}
\caption{Cosine similarity between the reference state and the state of each corresponding layer for \texttt{DINOV3}+Llama3.}
\label{fig:dionv3_llama}
\end{figure}

\begin{figure}[h]
\centering
\includegraphics[width=1.0\linewidth]{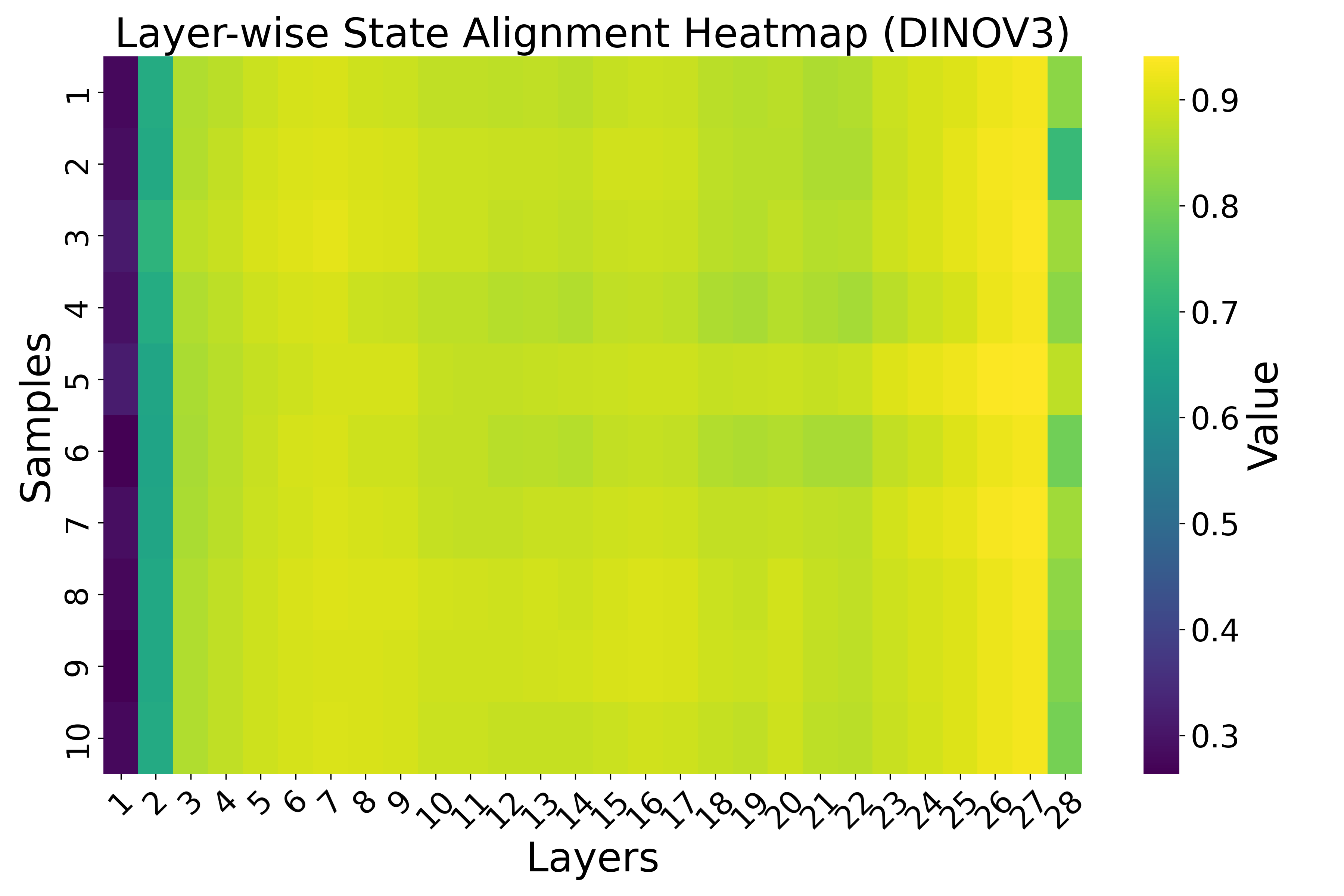}
\caption{Cosine similarity between the reference state and the state of each corresponding layer for \texttt{DINOV3}+Qwen2.5.}
\label{fig:dinov3_qwen}
\vspace{1.2 em}
\end{figure}

\begin{figure}[h]
\centering
\includegraphics[width=1.0\linewidth]{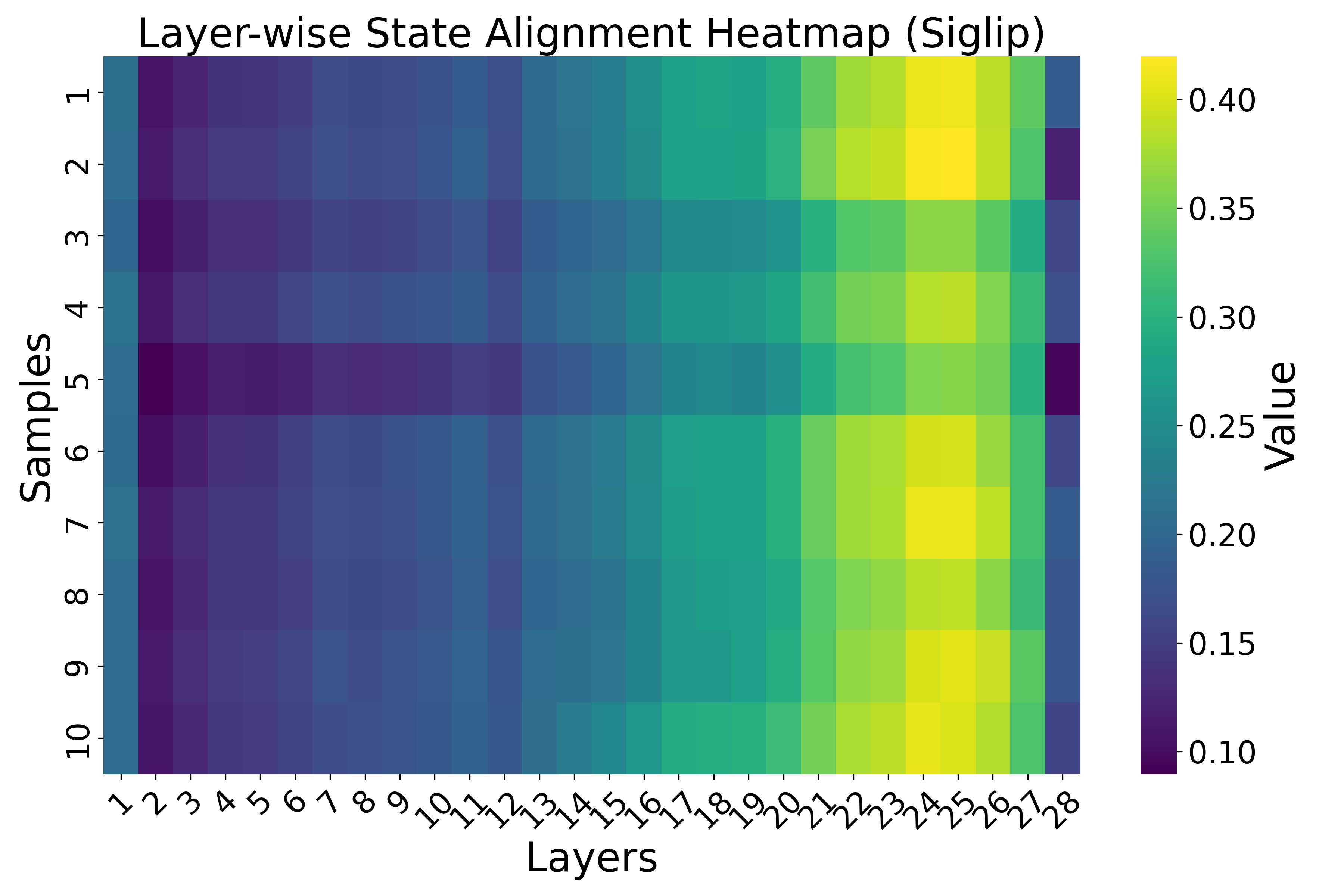}
\caption{Cosine similarity between the reference state and the state of each corresponding layer for \texttt{SigLIP}+Llama3.}
\label{fig:siglip_llama}
\end{figure}

\begin{figure}[h]
\centering
\includegraphics[width=1.0\linewidth]{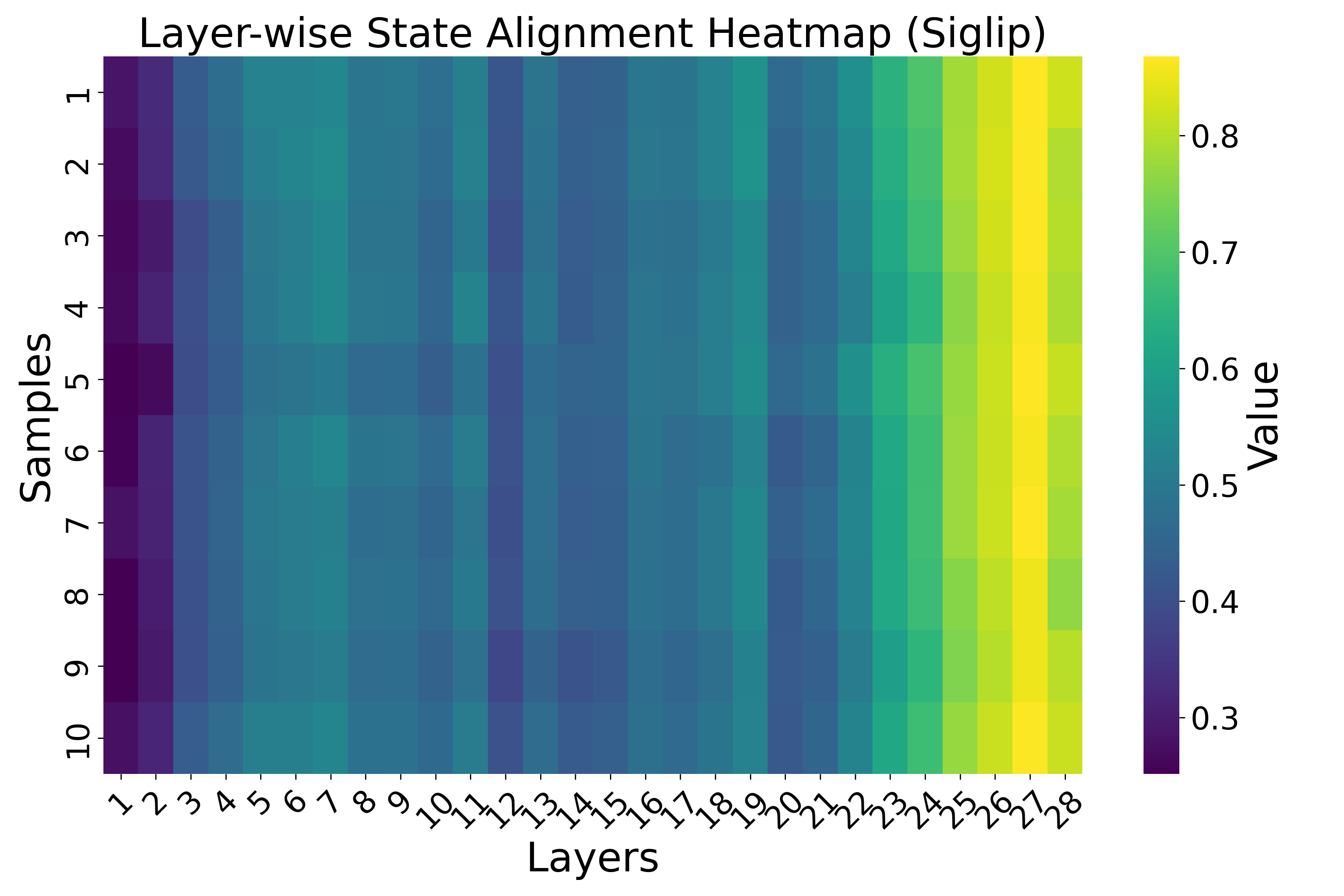}
\caption{Cosine similarity between the reference state and the state of each corresponding layer for \texttt{SigLIP}+Qwen2.5.}
\label{fig:siglip_qwen}
\end{figure}

\begin{figure}[h]
\centering
\includegraphics[width=1.0\linewidth]{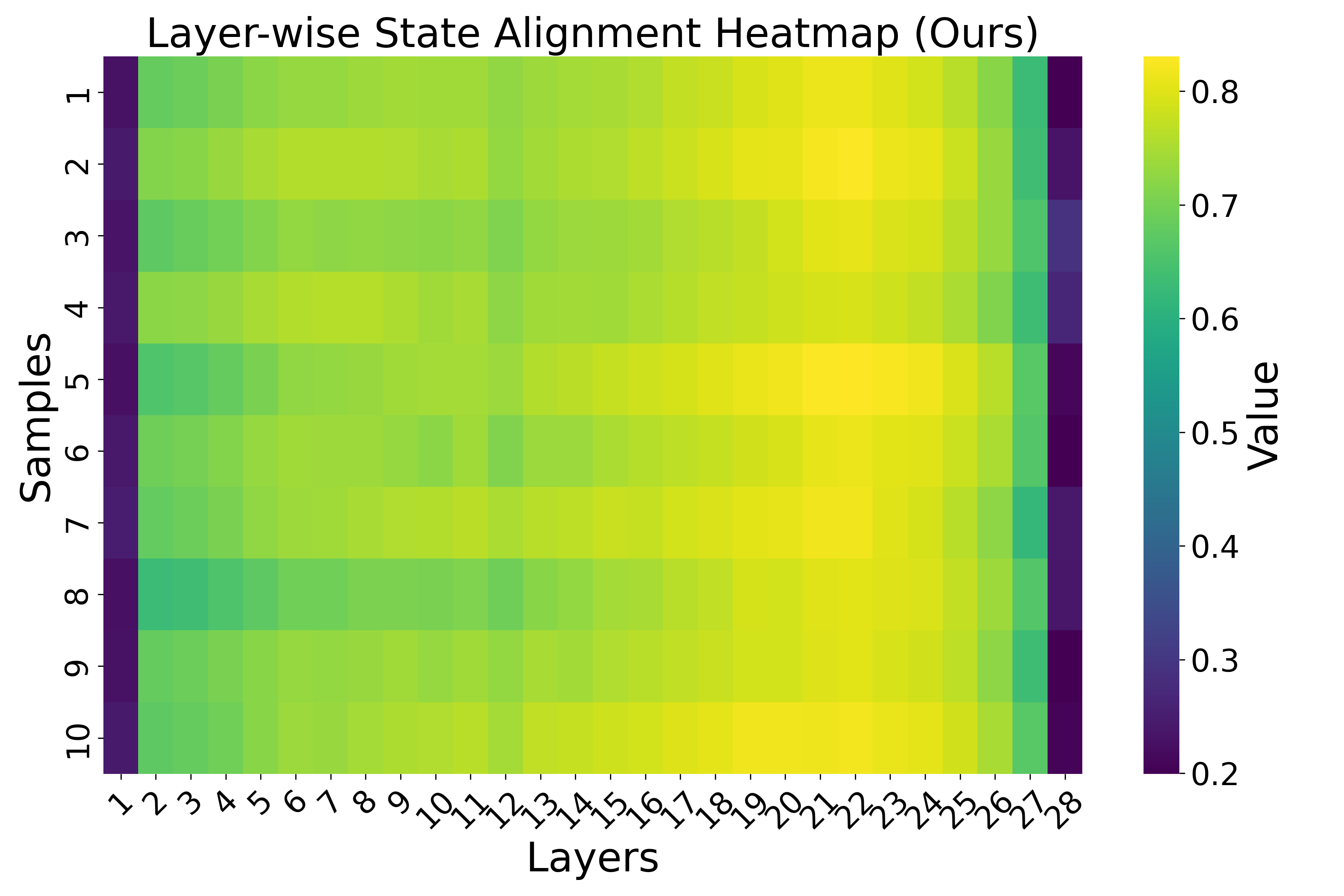}
\caption{Cosine similarity between the reference state and the state of each corresponding layer for \texttt{Granulon}+Llama3.}
\label{fig:ours_llama}
\vspace{1.2 em}
\end{figure}

\begin{figure}[h]
\centering
\includegraphics[width=1.0\linewidth]{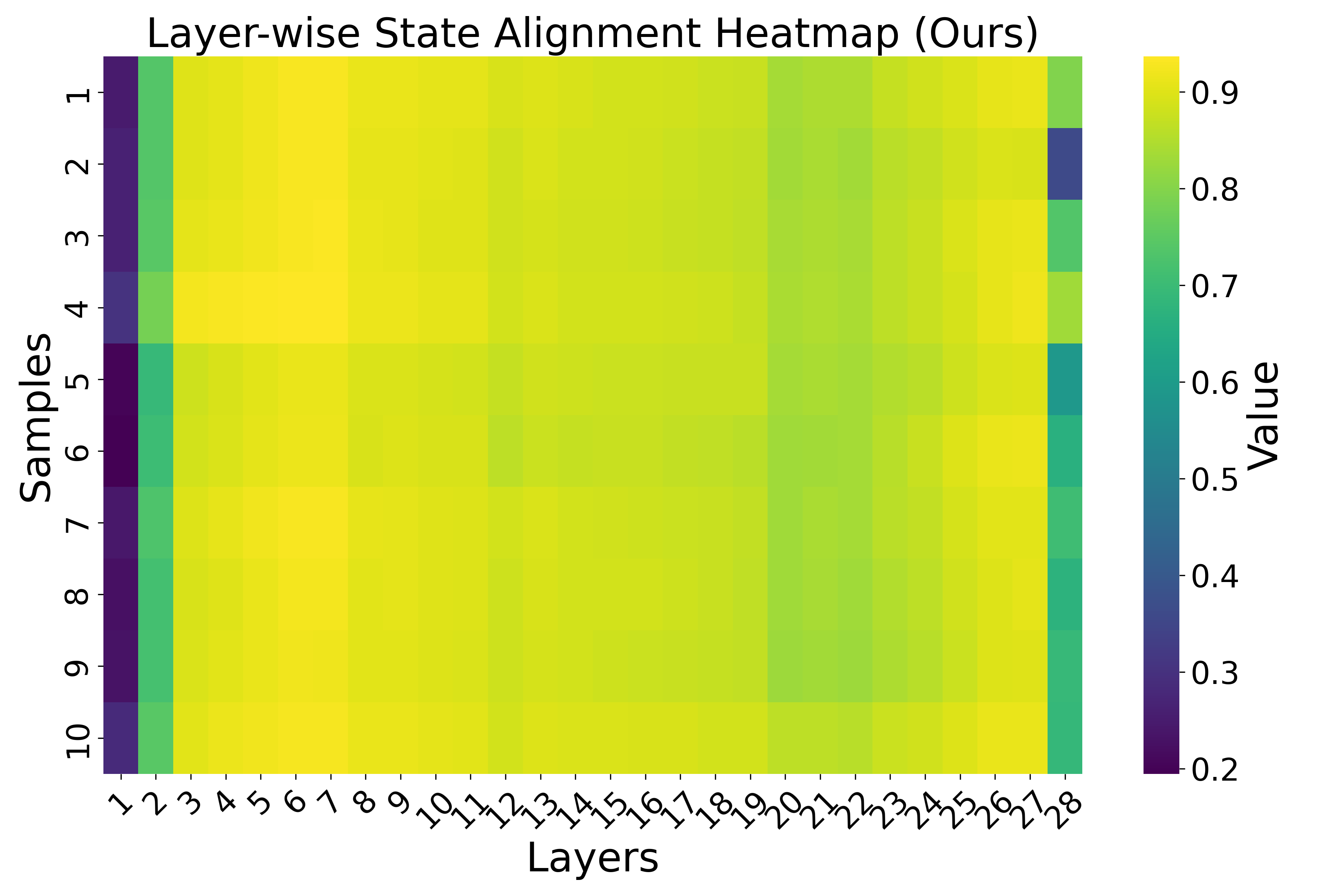}
\caption{Cosine similarity between the reference state and the state of each corresponding layer for \texttt{Granulon}+Qwen2.5.}
\label{fig:ours_qwen}
\end{figure}

We evaluate the layer-wise alignment of different visual encoders under two LLM backbones. As shown in Figures~\ref{fig:clip_llama}–\ref{fig:ours_qwen}, Granulon consistently achieves stronger cross-layer alignment than all baseline encoders across both backbones, demonstrating that our method effectively enhances internal representation alignment, semantic understanding, and reasoning capability.

\section{More Qualitative Case Study}
In this section, we provide more qualitative case studies to further illustrate the advantages of \texttt{Granulon}. Overall, \texttt{Granulon} achieves higher semantic accuracy, and its remaining errors tend to be confined to fine-grained visual details, likely arising from cases where the corresponding visual concepts are not sufficiently learned. In contrast, \texttt{CLIP}-based models exhibit lower accuracy and significantly higher hallucination rates. Moreover, once a hallucinated concept appears, \texttt{CLIP} often amplifies it, generating additional related but nonexistent details, leading to cascading semantic drift.

\vspace{2.0em}
\begin{tcolorbox}[colback=gray!10!white, colframe=gray!80!black, title=Case Study: Reasoning Answer, enhanced, breakable,]

\small\itshape   %
\setlength{\intextsep}{12pt}
\setlength{\columnsep}{10pt}
\begin{wrapfigure}{l}{0.35\linewidth} %
    \vspace{-1em}
    \includegraphics[width=\linewidth]{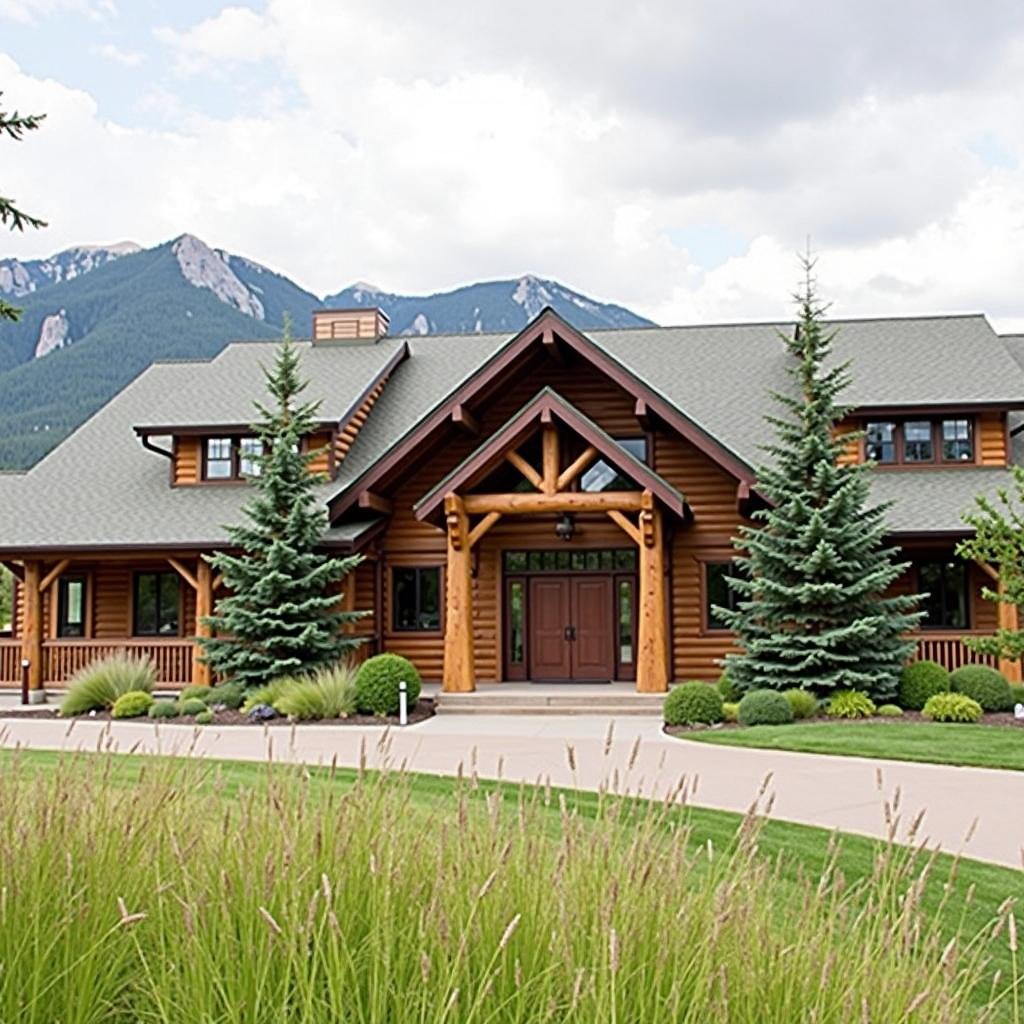}
    \vspace{-2em}
\end{wrapfigure}

\textbf{\texttt{CLIP}:} Set against a backdrop of serene mountainous terrain, this meticulously designed residential property exudes a sense of luxury and tranquility. The main entrance is framed by a \textcolor{green!50!black}{grand wooden gate}, adorned with a classic brass handle and a \textcolor{red!70!black}{welcoming inscription, ``Welcome Home,''} evoking a sense of warmth and hospitality. The gate leads into a \textcolor{green!50!black}{spacious driveway}, lined with \textcolor{red!70!black}{neatly trimmed hedges} and \textcolor{green!50!black}{a row of mature trees}, creating a welcoming path that leads to the front door. The front door itself is a focal point, featuring \textcolor{green!50!black}{a large, arched window}, adding a touch of elegance and sophistication. \textcolor{red!70!black}{The interior} is designed with \textcolor{red!70!black}{a blend of modern and rustic elements}, showcasing a combination of \textcolor{red!70!black}{hardwood floors}, \textcolor{red!70!black}{polished stone}, and \textcolor{red!70!black}{natural wood accents}, complemented by a spacious \textcolor{red!70!black}{living room, a dining area, and a gourmet kitchen} with stainless steel appliances. The property also features \textcolor{red!70!black}{a large deck with a pergola overhead}, perfect for outdoor entertaining and relaxation. The overall composition is harmonious, with the mountainous landscape providing a natural backdrop, enhancing the serene and peaceful atmosphere of this luxurious retreat.

\textbf{OURS:} 
Set against a backdrop of \textcolor{green!50!black}{pristine snow-covered mountains}, a \textcolor{green!50!black}{rustic wooden lodge} stands prominently, its distinctive design and warm tones evoking a sense of comfort and tranquility. The lodge, with its \textcolor{green!50!black}{peaked roof} and \textcolor{green!50!black}{overhanging eaves}, features a \textcolor{green!50!black}{central chimney} and \textcolor{green!50!black}{a pair of windows} on either side, adding to its inviting atmosphere. The exterior is adorned with a mix of natural elements, including a row of \textcolor{green!50!black}{neatly trimmed evergreen trees} and a cluster of colorful flowering \textcolor{green!50!black}{shrubs}, creating a harmonious blend of nature and craftsmanship. The surrounding landscape is dotted with small, well-maintained \textcolor{green!50!black}{gardens} and \textcolor{green!50!black}{lawns}, further emphasizing the lodge's peaceful setting.

\textbf{Ground Truth:}
Nestled amidst the breathtaking beauty of Jackson Hole, the Moose Creek Ranch Wedding Venue stands as a testament to rustic elegance and natural splendor. The image captures a majestic log cabin, its exterior crafted from rich, warm wood that exudes a sense of timeless charm and comfort. The structure features a steeply pitched roof with dark gray shingles, complemented by exposed wooden beams and a prominent central entrance framed by sturdy log columns. Flanking the doorway are two tall evergreen trees, their vibrant green foliage adding a touch of serenity and enhancing the venue's connection to nature. The front lawn is meticulously landscaped with neatly trimmed shrubs, lush grass, and a pathway leading up to the entrance, inviting guests into the tranquil setting. In the background, the majestic mountains rise against a partly cloudy sky, their rugged peaks providing a stunning backdrop that underscores the venue's remote and pristine location. The overall composition harmoniously blends the rugged beauty of the landscape with the refined details of the cabin, creating an idyllic setting where romance and adventure converge. The scene evokes a sense of peace and wonder, perfectly capturing the essence of a destination wedding venue that seamlessly integrates with the natural world.
\end{tcolorbox}

\vspace{-1.0em}
\begin{tcolorbox}[colback=gray!10!white, colframe=gray!80!black, title=Case Study: Reasoning Answer, enhanced, breakable, ]

\small\itshape   %
\setlength{\intextsep}{12pt}
\setlength{\columnsep}{10pt}
\begin{wrapfigure}{l}{0.35\linewidth} %
    \vspace{-1em}
    \includegraphics[width=\linewidth]{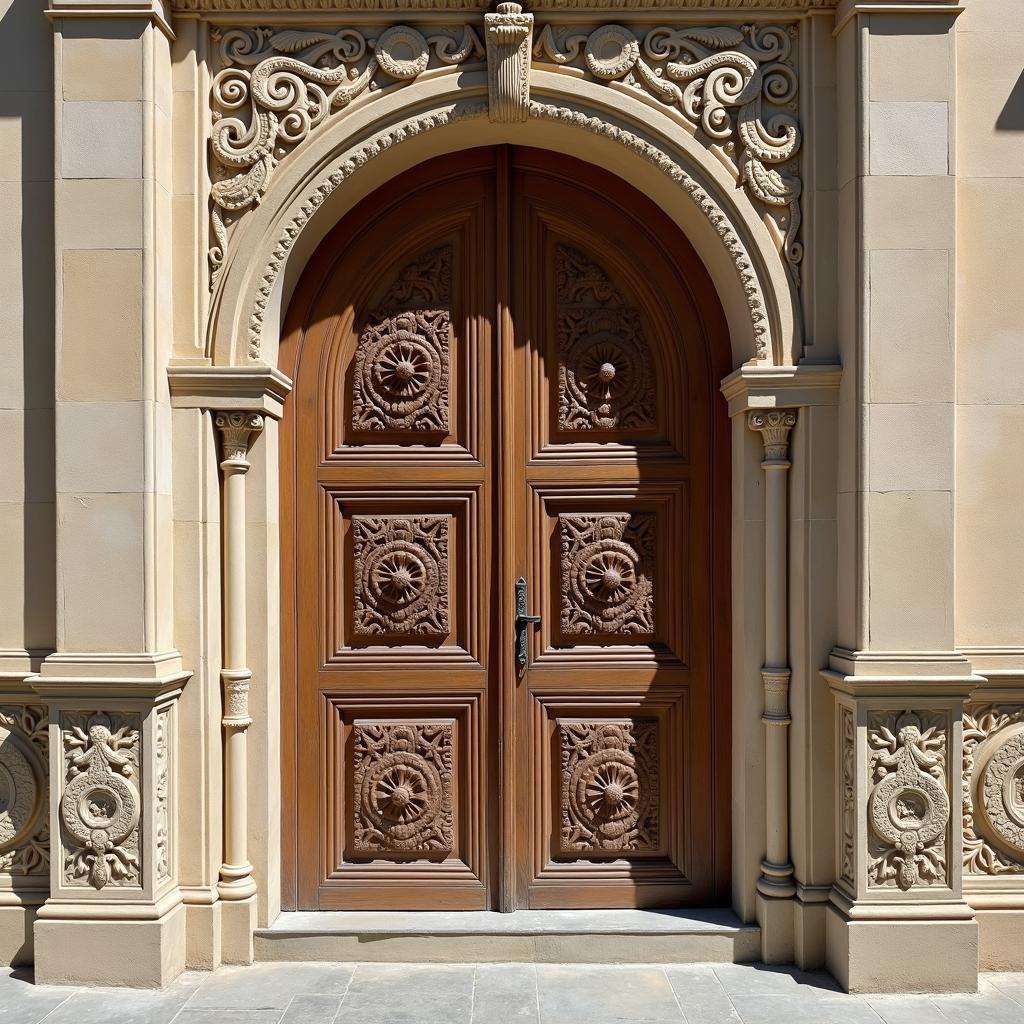}
    \vspace{-2em}
\end{wrapfigure}

\textbf{\texttt{CLIP}:} Elegant and opulent, the image depicts a \textcolor{red!70!black}{grand staircase} with intricate architectural details, showcasing a blend of classical and modern design elements. The central staircase is adorned with ornate ironwork, featuring \textcolor{red!70!black}{a series of ornate columns} and capitals \textcolor{red!70!black}{that rise gracefully from the floor}, each adorned with intricate carvings and decorative motifs. The staircase is flanked by \textcolor{red!70!black}{two pairs of columns, with each pair supporting a balustrade that extends upward, creating a sense of height and grandeur}. The balustrades are adorned with delicate ironwork, adding a touch of elegance and symmetry to the design. At the top of the staircase, \textcolor{red!70!black}{a large arched window with a wrought iron frame and ornate ironwork above it} adds a touch of architectural flair. \textcolor{red!70!black}{The floor beneath the staircase is covered in a dark wood}, enhancing the overall opulence and symmetry of the space. The composition is balanced and harmonious, with the staircase serving as the focal point, while the surrounding elements complement and enhance the overall aesthetic. The image captures a moment of grandeur, evoking a sense of timeless beauty and sophistication.

\textbf{OURS:} 
Question: What is the architectural style of the building in the image?$\backslash n\backslash n$Answer: The building in the image is a \textcolor{green!50!black}{classic example of Baroque architecture}, characterized by its ornate and symmetrical design, particularly evident in the \textcolor{green!50!black}{intricate stone carvings and gilded details}. The façade features a grand entrance with a large, arched doorway flanked by two smaller, \textcolor{red!70!black}{arched windows}, creating a sense of balance and harmony. The doorway is adorned with a \textcolor{green!50!black}{decorative keystone}, and the surrounding stonework is detailed with \textcolor{green!50!black}{interlocking arches and ornate patterns}, suggesting a high level of craftsmanship. The overall composition is symmetrical, with the \textcolor{green!50!black}{central doorway mirroring the symmetry of the surrounding elements}, reinforcing the building's grandeur and classical aesthetic. The image captures the building's ornate details, showcasing the Baroque style's emphasis on grandeur, symmetry, and intricate ornamentation.

\textbf{Ground Truth:}
Elegantly carved wooden double doors, crafted from rich, polished wood, form the centerpiece of this grand architectural entrance, showcasing intricate floral and geometric patterns that adorn each panel. The doors are framed by an ornate stone archway, intricately sculpted with swirling motifs and classical detailing, which adds a sense of timeless elegance and sophistication. Flanking the doorway are decorative columns, each embellished with bas-reliefs featuring elaborate floral designs and symbolic motifs, further enhancing the historical and artistic significance of the structure. The stone facade surrounding the entrance is composed of light beige blocks, providing a neutral backdrop that allows the detailed carvings on both the doors and the arch to stand out prominently. The overall composition is symmetrical and balanced, with the central focus on the arched doorway, drawing the viewer's eye towards its intricate craftsmanship. The scene exudes a sense of reverence and history, suggesting that this is the entrance to a historic church or a similarly significant building, where every element has been meticulously designed to convey beauty, tradition, and spiritual grandeur. The lighting highlights the textures and shadows, emphasizing the three-dimensional quality of the carvings and adding depth to the image. The absence of any visible text within the image itself ensures that the visual narrative remains unbroken, allowing the architectural details to speak for themselves. The style is photorealistic, capturing the fine details and textures with precision.

\end{tcolorbox}

\vspace{-1.0em}
\begin{tcolorbox}[colback=gray!10!white, colframe=gray!80!black, title=Case Study: Reasoning Answer, enhanced, breakable,]

\small\itshape   %
\setlength{\intextsep}{12pt}
\setlength{\columnsep}{10pt}
\begin{wrapfigure}{l}{0.35\linewidth} %
    \vspace{-1em}
    \includegraphics[width=\linewidth]{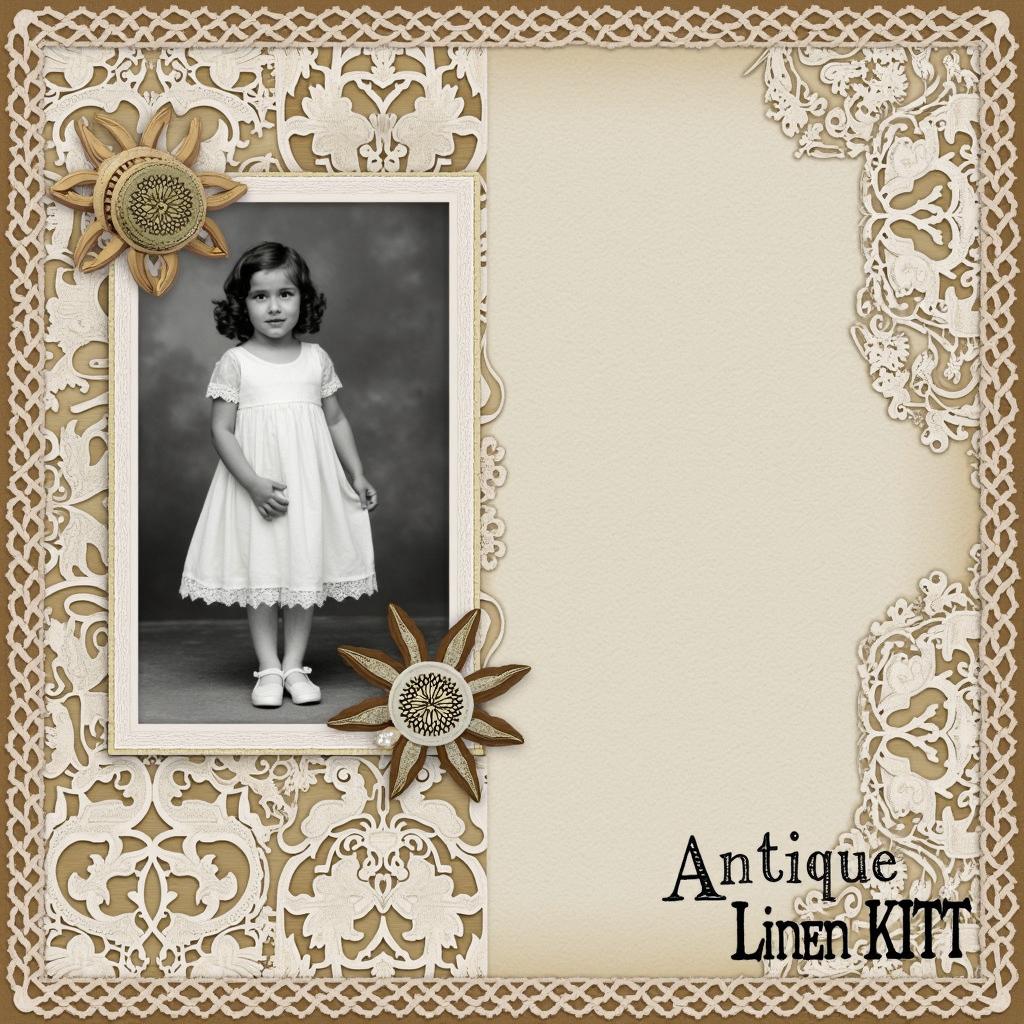}
    \vspace{-2em}
\end{wrapfigure}

\textbf{\texttt{CLIP}:} Elegant and timeless, this vintage-style illustration captures the essence of a traditional school uniform, set against a soft, pastel background. The composition features a young girl \textcolor{red!70!black}{standing in front of a classroom, her attire perfectly aligned with the school's uniform standards}. Her attire includes a \textcolor{red!70!black}{navy blue blouse with a white collar, complemented by a matching navy blue skirt}, both adorned with subtle white embroidery. The uniform is further enhanced by a white apron, which adds a touch of elegance and practicality. The girl's hair is styled in a neat bun, with a small bow at the back, completing her formal appearance. \textcolor{red!70!black}{The classroom setting is depicted with a clean, white chalkboard and a wooden desk}, symbolizing the educational environment. The overall style is classic and refined, evoking a sense of tradition and respect for academic pursuits. The image is rendered in a detailed, hand-drawn style, capturing the intricate details of the uniform and the girl's pose with precision. The composition is balanced and harmonious, emphasizing the girl's focus and the solemnity of the \textcolor{red!70!black}{educational setting}. The use of pastel colors and the overall serene atmosphere create a warm and inviting backdrop for the uniform, highlighting its timeless appeal.

\textbf{OURS:} 
The primary purpose of the image is to showcase a beautiful, high-quality photograph of a \textcolor{green!50!black}{young girl dressed in a white, lace-trimmed dress} with a matching \textcolor{red!70!black}{hat}, standing in a serene and elegant pose. The composition is centered, with the girl's face slightly tilted upwards, exuding a sense of innocence and charm. \textcolor{green!50!black}{The background is a soft, creamy white}, providing a clean and timeless backdrop for the subject. The image is framed by a \textcolor{green!50!black}{decorative border featuring floral patterns and a subtle gold trim}, adding a touch of sophistication and elegance. The overall style is reminiscent of classic, vintage photography, with a focus on capturing the beauty and innocence of the subject.

\textbf{Ground Truth:}
Elegantly composed within an intricately designed lace border, a vintage black-and-white photograph captures the timeless charm of a young girl standing gracefully in a simple yet sophisticated white dress. The dress, adorned with delicate lace trim at the hem, complements her poised stance, hands gently clasped in front. The photograph is framed by an ornate lace pattern that extends across the edges, adding a layer of nostalgic sophistication. Two decorative sunflower embellishments, one positioned above and another below the portrait, enhance the vintage aesthetic with their warm, golden tones and intricate detailing. The textured background, labeled \"Antique Linen KITT,\" further enriches the composition, suggesting a connection to Gabby~Swirlydoo's Antique Linen Kit, which inspires this nostalgic and elegant design. The overall style is a harmonious blend of vintage-inspired elements, seamlessly integrating classic photographic portraiture with artisanal lacework and decorative accents, evoking a serene and timeless atmosphere that bridges the past and present.

\end{tcolorbox}

\begin{tcolorbox}[colback=gray!10!white, colframe=gray!80!black, title=Case Study: Reasoning Answer, enhanced, breakable,]

\small\itshape   %
\setlength{\intextsep}{12pt}
\setlength{\columnsep}{10pt}
\begin{wrapfigure}{l}{0.35\linewidth} %
    \vspace{-1em}
    \includegraphics[width=\linewidth]{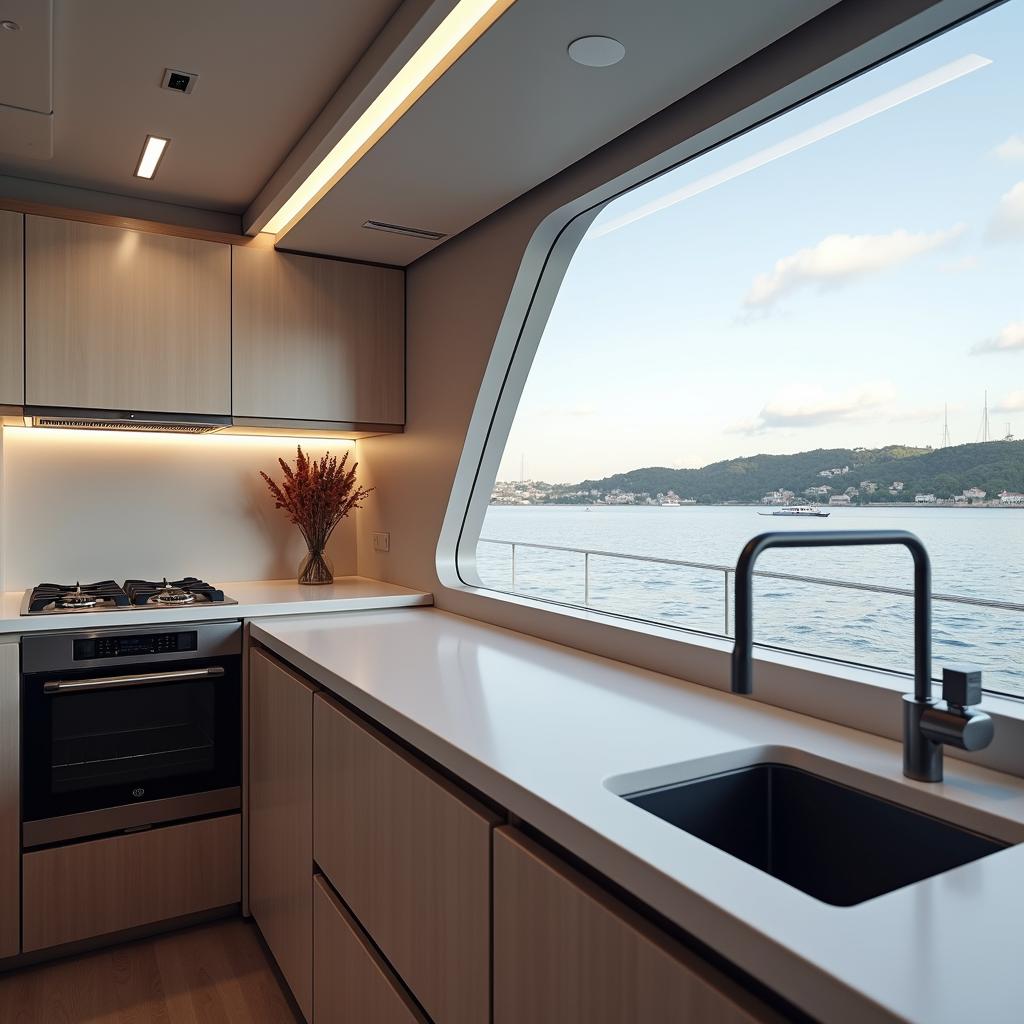}
    \vspace{-2em}
\end{wrapfigure}

\textbf{\texttt{CLIP}:} Sleek and modern, the image depicts a \textcolor{green!50!black}{minimalist yet elegant interior of a yacht}, showcasing a harmonious blend of natural light and sophisticated design. The vessel, likely a luxury yacht, is anchored in a calm, blue-green \textcolor{green!50!black}{lagoon}, its sleek hull glinting under the soft glow of the setting sun. The interior is bathed in a warm, inviting atmosphere, with a minimalist palette of \textcolor{green!50!black}{white walls and wooden accents} that create a serene and spacious feel. \textcolor{red!70!black}{The yacht's spacious deck} features a \textcolor{red!70!black}{large, open-air living area with a circular table and chairs}, inviting guests to gather and enjoy meals or conversations. \textcolor{red!70!black}{The table is adorned with a simple, elegant centerpiece, likely a vase or decorative item}, adding a touch of elegance. The walls are painted in a soft, gradient blue hue, complementing the natural light filtering through the open windows, which frame the expansive view of the lagoon. The yacht's interior is designed to maximize natural light, with \textcolor{green!50!black}{large windows} and skylights that allow for ample illumination, enhancing the overall sense of tranquility and luxury. The composition is balanced, with the yacht's sleek lines and the serene lagoon providing a backdrop of calm and beauty, making it a perfect representation of a luxurious and serene maritime experience.

\textbf{OURS:} 
$\backslash n$A) To provide a functional space for food preparation and cooking.$\backslash n$B) To serve as a dining area for guests.$\backslash n$C) To showcase a luxurious and modern aesthetic.$\backslash n$D) To offer a serene and peaceful environment for relaxation.$\backslash n\backslash n$Analysis:$\backslash n$The image depicts a \textcolor{green!50!black}{sleek and modern kitchen}, characterized by its minimalist design and emphasis on functionality. The primary function of this kitchen is evident in its well-organized layout, which prioritizes efficiency and ease of use. The white \textcolor{green!50!black}{cabinets} and countertops create a clean and contemporary look, while the stainless \textcolor{green!50!black}{steel sink and faucet} add a touch of industrial chic. The kitchen island, positioned at the center, serves as a \textcolor{green!50!black}{central hub} for food preparation and cooking, featuring a sleek granite countertop and a built-in stove. The overall composition is harmonious, with the kitchen island and countertops aligned to create a sense of balance and order. The inclusion of a \textcolor{green!50!black}{window} above the sink offers a stunning view of the ocean, enhancing the kitchen's connection to nature and providing a calming backdrop for cooking and dining. The lighting is soft and natural, with a pendant light above the kitchen island adding warmth and ambiance. The image exudes a sense of modernity and functionality.

\textbf{Ground Truth:}
Modern yacht kitchen within the 2018 Absolute 58' NAVETTA, showcasing a sleek and minimalist design with warm wooden accents and high-definition detail. The space features sleek white countertops paired with a black sink and a minimalist faucet, complemented by under-cabinet lighting that enhances the warm ambiance. A large panoramic window dominates the right side, offering a serene view of a waterfront landscape with distant hills, a calm body of water, and a small boat gently floating on the surface. The kitchen includes a modern gas stove with four burners and a built-in oven below, both integrated seamlessly into the cabinetry. A vase of dried flowers sits atop the counter, adding a touch of natural elegance. The ceiling features linear LED lighting strips and recessed fixtures, providing ample illumination while maintaining a clean, contemporary aesthetic. The overall composition emphasizes natural light and contemporary design, capturing the essence of luxury and tranquility aboard this sophisticated vessel.

\end{tcolorbox}

\begin{tcolorbox}[colback=gray!10!white, colframe=gray!80!black, title=Case Study: Reasoning Answer, enhanced, breakable,]

\small\itshape   %
\setlength{\intextsep}{12pt}
\setlength{\columnsep}{10pt}
\begin{wrapfigure}{l}{0.35\linewidth} %
    \vspace{-1em}
    \includegraphics[width=\linewidth]{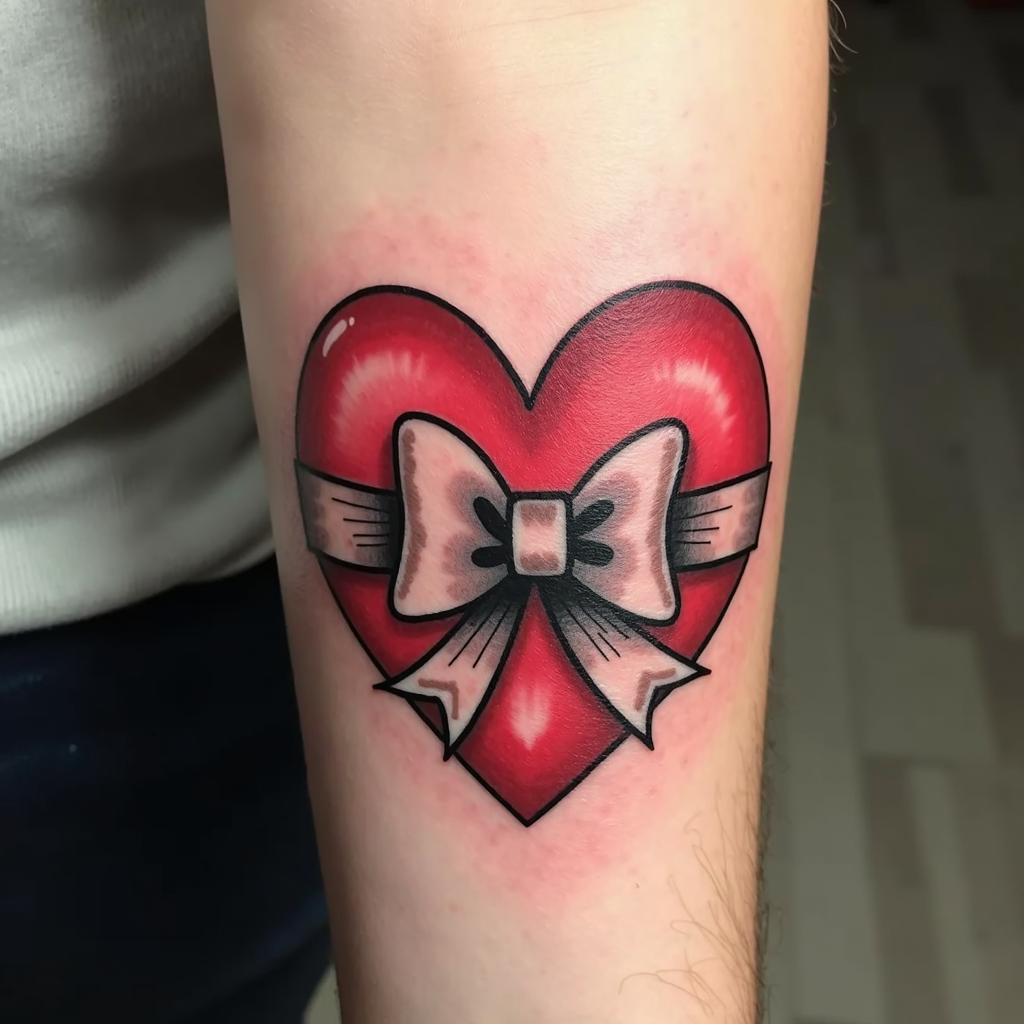}
    \vspace{-2em}
\end{wrapfigure}

\textbf{\texttt{CLIP}:} Centered on a \textcolor{red!70!black}{pristine white background}, a striking black and \textcolor{red!70!black}{gold} tattoo design captures the essence of a vibrant and intricate piece of art. The tattoo features a \textcolor{red!70!black}{bold, circular pattern that radiates outward from the center, with a central black dot surrounded by a series of concentric rings in gold}, each one slightly larger than the last. The design is rendered in a high-contrast style, with the black dot and rings standing out sharply against the white background, creating a dynamic and eye-catching visual effect. The tattoo's composition is \textcolor{green!50!black}{balanced and symmetrical}, with the \textcolor{red!70!black}{central dot and rings} forming a harmonious whole, suggesting a sense of unity and depth. The overall style is highly stylized and graphic, reminiscent of modern tattoo art, characterized by its bold lines, vibrant colors, and precise execution. The image exudes a sense of creativity and artistic expression, making it a captivating piece of wearable art.

\textbf{OURS:} 
Question: What is the significance of the red heart symbol in the image?$\backslash n\backslash n$Analysis: The image features \textcolor{green!50!black}{a close-up of a hand-drawn heart tattoo on a person's hand, with the red heart symbol prominently displayed at its center}. The tattoo is rendered in a simple yet elegant style, with \textcolor{green!50!black}{bold lines and a vibrant red color that gives it a striking contrast against the darker background}. The heart is surrounded by a delicate chain, which adds a touch of whimsy and playfulness to the design. The overall composition is centered, drawing the viewer's attention directly to the heart, emphasizing its central role in the tattoo. The background is blurred, focusing the viewer's focus on the tattoo itself, while the hand provides a natural frame that grounds the image in a realistic context. The image exudes a sense of intimacy and personal expression, suggesting that the tattoo is a meaningful symbol of love or commitment.

\textbf{Ground Truth:}
Vibrantly displayed on the inner forearm is a striking heart tattoo, rendered in bold red with intricate detailing that brings it to life. The heart shape is filled with a glossy, three-dimensional texture, highlighted by careful shading and light reflections, giving it a lifelike appearance. At its center sits a delicate pink bow, tied neatly with a black ribbon that loops gracefully around the heart's contours. The bow features subtle gradients and shadows, enhancing its realism and adding depth to the overall design. The tattoo's crisp lines and precise execution suggest a clean, vector-style illustration technique, blending modern aesthetics with traditional tattoo artistry. The surrounding skin appears smooth and lightly textured, with some natural hair visible near the bottom edge, grounding the vibrant artwork within a realistic context. The composition is centered and balanced, drawing immediate attention to the heart and bow, symbolizing love and elegance in a bold yet refined manner.

\end{tcolorbox}